\documentclass[10pt,journal,compsoc]{IEEEtran}

\usepackage{booktabs} 
\usepackage{tabularx}
\usepackage{graphicx}
\usepackage{url}
\usepackage{mdwlist}
\usepackage{subcaption}
\usepackage{hyperref}
\hypersetup{colorlinks=false,linkcolor=blue,urlcolor=blue,citecolor=red}
\usepackage{amsfonts}
\usepackage{amsmath}
\usepackage[labelfont=bf,textfont=bf]{caption}
\usepackage{multirow}
\usepackage{array}
\usepackage[linesnumbered,ruled]{algorithm2e}
\SetKwComment{Comment}{$\triangleright$\ }{}
\usepackage{xcolor}
\usepackage{amssymb}
\usepackage[flushleft]{threeparttable}
\usepackage{enumitem}
\usepackage{balance}
\usepackage{algorithm2e}

\newcommand{\ie}{\textit{i.e.}}
\newcommand{\eg}{\textit{e.g.}}
\newcommand{\etal}{\textit{et al.}}

\DeclareMathOperator*{\argmax}{argmax}
\graphicspath{ {./figures/} }

\ifCLASSOPTIONcompsoc
  \usepackage[nocompress]{cite}
\else
  \usepackage{cite}
\fi

\begin{document}

\title{Learning to Compose and Reason with Language Tree Structures for Visual Grounding}
\author{Richang~Hong, Daqing~Liu, Xiaoyu~Mo, Xiangnan~He, Hanwang~Zhang
\IEEEcompsocitemizethanks{\IEEEcompsocthanksitem R. Hong is with the School of Computer \& Information, Hefei University of Technology, Hefei, China. E-mail: see https://sites.google.com/site/homeofrichanghong/
\IEEEcompsocthanksitem D. Liu is with the University of Science and Technology of China, Hefei, China. E-mail: liudq@mail.ustc.edu.cn
\IEEEcompsocthanksitem X. Mo is with the School of Electrical \& Electronic Engineering, Nanyang Technological University, Singapore. E-mail: moxy@ntu.edu.sg
\IEEEcompsocthanksitem X. He is with the School of Information Science and Technology, University of Science and Technology of China. E-mail: xiangnanhe@gmail.com
\IEEEcompsocthanksitem H. Zhang is with the School of Computer Science \& Engineering, Nanyang Technological University, Singapore. E-mail: see http://www.ntu.edu.sg/home/hanwangzhang/
}
}

\markboth{IEEE Transactions on Pattern Recognition and Machine Intelligence}%
{Hong \MakeLowercase{\textit{et al.}}: Learning to Compose and Reason with Language Tree Structures for Visual Grounding}

\IEEEtitleabstractindextext{%
\begin{abstract}
Grounding natural language in images, such as localizing ``the black dog on the left of the tree'', is one of the core problems in artificial intelligence, as it needs to comprehend the fine-grained language compositions. However, existing solutions merely rely on the association between the holistic language features and visual features, while neglect the nature of composite reasoning implied in the language. In this paper, we propose a natural language grounding model that can automatically compose a binary tree structure for parsing the language and then perform visual reasoning along the tree in a bottom-up fashion. We call our model \textsc{RvG-Tree}: Recursive Grounding Tree, which is inspired by the intuition that any language expression can be recursively decomposed into two constituent parts, and the grounding confidence score can be recursively accumulated by calculating their grounding scores returned by the two sub-trees. \textsc{RvG-Tree} can be trained end-to-end by using the Straight-Through Gumbel-Softmax estimator that allows the gradients from the continuous score functions passing through the discrete tree construction. Experiments on several benchmarks show that our model achieves the state-of-the-art performance with more explainable reasoning.
\end{abstract}
}
\maketitle
\IEEEdisplaynontitleabstractindextext
\IEEEpeerreviewmaketitle

\section{Introduction}
\label{sec:introduction}
With the maturity of deep neural networks for object detection~\cite{redmon2017yolo9000}, we are more ambitious to fulfill the long-term goal in computer vision: an intelligent agent that can comprehend human instructions in natural language and execute them in visual environment. Once achieved, it will benefit various human-computer interaction applications such as visual Q\&A~\cite{antol2015vqa}, visual dialog~\cite{das2017learning}, and robotic navigation~\cite{Das_2018_CVPR}. To achieve this, a necessary step is to extend the current object detection system from fixed-sized inventory of words to open-vocabulary sentences, that is, grounding natural language in images~\cite{mao2016generation}.

\begin{figure}[ht]
\begin{subfigure}{.25\textwidth}
  \centering
  \includegraphics[width=.8\linewidth]{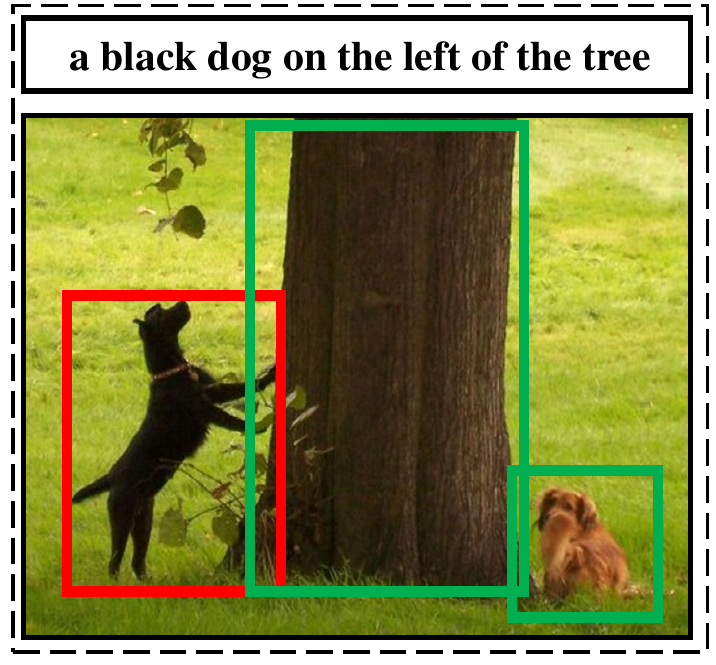}
  \caption{}
  \label{fig:1a}
\end{subfigure}%
\begin{subfigure}{.25\textwidth}
  \centering
  \includegraphics[width=.8\linewidth]{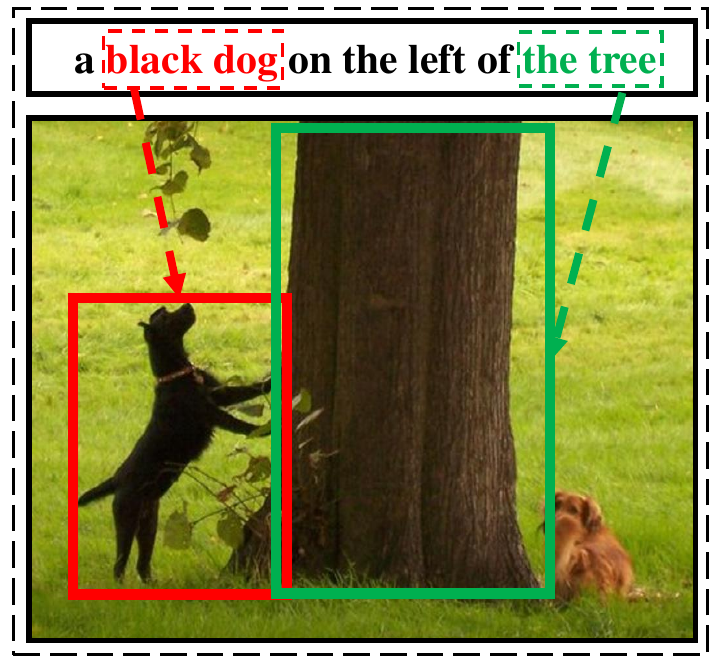}
  \caption{}
  \label{fig:1b}
\end{subfigure}
\begin{subfigure}{.5\textwidth}
  \centering
  \includegraphics[width=0.8\linewidth]{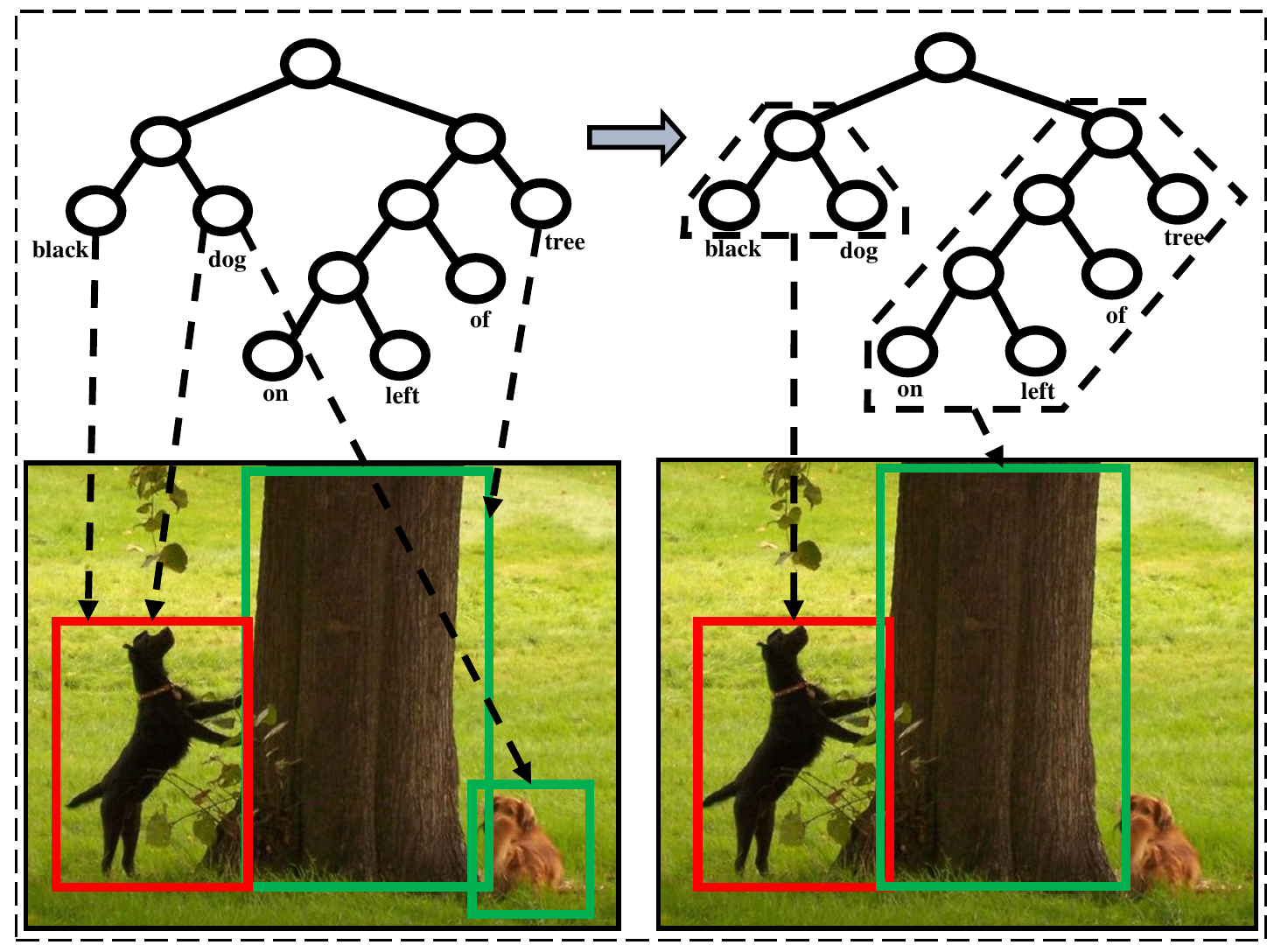}
  \caption{}
  \label{fig:1c}
\end{subfigure}
\vspace{-2mm}
\caption{ (a) A typical task of grounding the natural language ``a black dog on the left of the tree'' in the image, represented as a set of object bounding boxes. The output should be the ``dog'' grounded with the red box. (b) Most recent advances focus on simple compositions of the language such as (subject, predicate, object) triplet. (c) Our \textsc{RvG-Tree} decomposes the sentence into more fine-grained compositions, and then accumulates the grounding confidence score in a bottom-up fashion. }
\vspace{-4mm}
\label{fig:intro}
\end{figure}

Thanks to the advance of visual deep features~\cite{he2016deep} and neural language models~\cite{mikolov2010recurrent}, recent studies show promising results on scaling up visual grounding to open-vocabulary scenario, such as thousands of object categories~\cite{redmon2017yolo9000,hu2018learning}, relationships~\cite{zhang2017visual}, and phrases~\cite{plummer2017phrase}. However, grounding natural language (cf. Fig.~\ref{fig:1a}) is still far from satisfactory as the key is not only to associate related semantics to the target visual object, but also to distinguish it from the contextual objects, especially those of the same category. For example, as shown in Fig.~\ref{fig:1a}, to ground the referring expression ``a black dog on the left of the tree'', we need to first detect objects in the image and then distinguish the referent ``black dog'' from the other ones especially those with the same category ``golden dog'' using the context ``black'' and ``left of the tree''.

To make a successful discrimination between context and referent, we need to parse the language into corresponding semantic components. As illustrated in Fig.~\ref{fig:1b}, current state-of-the-art models~\cite{hu2017modeling,zhang2018grounding,yu2018mattnet} learn to parse a sentence into the (subject, predicate, object) triplets, and the referent grounding score is the sum of the three grounding scores. The intuition behind these compositional methods is that the parsing helps to divide the original problem into easier sub-tasks, \ie, finding the contextual regions that grounded by ``predicate'' and ``object'' semantics is apparently helpful to localize the referent. However, we argue that the above triplet composition for a sentence is still too coarse. For example, it is meaningful to parse short sentences such as ``person riding bike'' into triplets, as it has a clear grounding for individual  ``person'', ``bike'', and their relationship; but it is problematic for general longer sentences with adjective clause, \eg, it is still difficult to parse the following long sentence into one triplet: ``a black dog on the left of the tree which is bigger than others''.

In this paper, we propose a fine-grained natural language grounding model called \textbf{R}ecursi\textbf{v}e \textbf{G}rounding \textbf{Tree} (\textsc{RvG-Tree}). The key motivation is to decompose any language sentence into semantic constituents in a recursive way, that is, every object has a clause modifier, which can be further parsed into its own object and modifier clause. As illustrated in Fig.~\ref{fig:1c}, ``black dog'' can be decompose into ``black'' and ``dog'', and thus the compositional confidence for ``black dog'' can be accumulated by ``something is black'' and ``something is a dog''; and the rest ``on the left of the tree'' can be further decomposed into ``on the left'' and ``tree'', and it helps to localize ``something is on the left of the tree''. Therefore, by using \textsc{RvG-Tree}, we can accumulate the grounding confidence score from the lower layers which are relatively simpler grounding sub-tasks. Compared to previous methods that rely on sentence embedding features, \textsc{RvG-Tree} offers an explainable way of understanding how the language is comprehended in visual grounding. It is worth noting that not all the nodes of \textsc{RvG-Tree} contribute to the final score. In particular, we design a classifier that determines whether a node should return a visual feature or a score, where the former is used as the contextual feature for the higher-level, and the latter is used for score accumulation. Thanks to this design, our \textsc{RvG-Tree} is generic and flexible and thus can be applied in longer natural language sentences.

\begin{figure*}
\centering
\includegraphics[width=0.98\textwidth, angle=0]{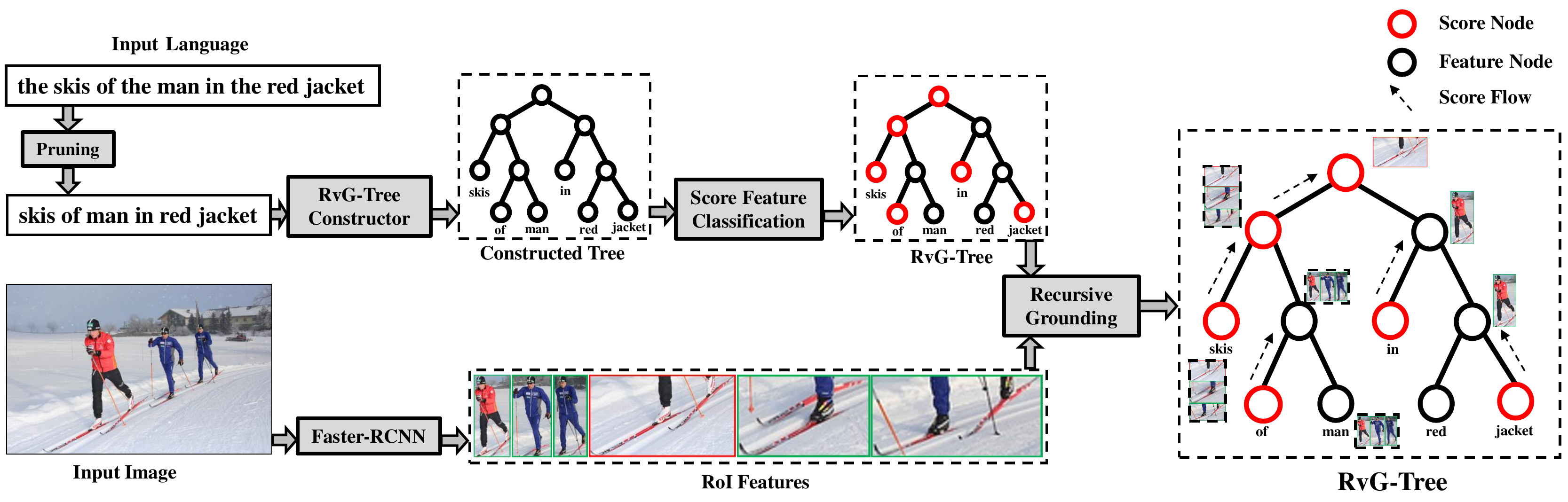}
\vspace{-2mm}
\caption{ The overview of using \textsc{RvG-Tree} for natural language grounding. Given a natural language sentence as input, we first prune the sentence and then construct \textsc{RvG-Tree} (Section~\ref{sec:treeconstruction}). Then, we have a score-feature classifier~(Section~\ref{sec:recursivegrounding}) to determine each node as the ``score node'' or ``feature node'', where the score node returns the recursive score and the feature node returns the feature (Section~\ref{sec:recursivegrounding}). The final score of the root node is accumulated recursively in a bottom-up fashion (Section~\ref{sec:recursivegrounding}) and the visual region with the highest score is considered as the grounding result. Note that all the nodes can be visualized by the corresponding confidence scores and only qualitative regions are visualized.}
\vspace{-4mm}
\label{fig:2}
\end{figure*}

The technical overview of \textsc{RvG-Tree} is illustrated in Fig.~\ref{fig:2}. Inspired by the recent progress on tree structure construction for sentence representations~\cite{choi2018learning}, we propose to learn \textsc{RvG-Tree} in a bottom-up fashion, by dynamically merging any two adjacent nodes. Specifically, we start from leaf nodes which are words, where the two merged nodes are chosen based on their association score (\eg, ``black'' and ``dog''). Then, the merged and un-merged nodes are flushed to the next merging layer. Finally, the construction is complete when there are only one node left in the pool. Given an \textsc{RvG-Tree} constructed from a sentence, we design a recursive grounding score function that accumulates the grounding confidence from leaf to root. Considering any sub-tree with one root and two children nodes, we first use the node classifier to determine which children node is the \emph{score} or \emph{feature} node; the score node returns the grounding score from its own sub-tree, and the feature node returns the soft-attention weighted sum of the visual regions, where the weights are softmax-normalized grounding scores at this node. The overall grounding score contains two non-differentiable decision-making processes: 1) the node merging process --- choosing the highest association score in the pool --- in the \textsc{RvG-Tree} construction, and 2) the score and feature node classification in the recursive grounding score calculation. To this end, we use Gumbel-Softmax~\cite{gumbelsoftmax} with proper expert supervision to make the overall architecture fully-differentiable, \ie, standard SGD can be applied in the discrete decisions.

We perform extensive experiments on three challenging referring expression grounding datasets: RefCOCO~\cite{yu2016modeling}, RefCOCO+~\cite{yu2016modeling}, and RefCOCOg~\cite{mao2016generation}. Compared to existing grounding models, \textsc{RvG-Tree} is the first model that has totally transparent visual reasoning process for grounding and achieves comparative or even better performances.

Our contributions are summarized as follows:
\begin{itemize}[leftmargin=.1in]
	\item We propose \textsc{RvG-Tree}: a fine-grained vision-language reasoning model for visual grounding. 
	\item \textsc{RvG-Tree} introduces a novel tree structure to parse the language input and calculates the grounding score in an efficiently recursive fashion, allowing machines to understand natural language in a way similar to the language constitution.
	\item \textsc{RvG-Tree} is designed to be fully-differentiable and thus it can be trained efficiently with standard SGD. 
\end{itemize}

\section{Related Work}
\subsection{Grounding Natural Language}
Referring expressions are natural language sentences describing the referent objects within a particular scene, \eg, ``the man on the left of the golden dog'' or ``the dog on the sofa''. Grounding referring expression, which aims to localize the referring expression in a image, is also known as referring expression comprehension, and its inverse task is called referring expression generation \cite{mao2016generation}. Based on valid phrase grounding methods, referring expression grounding steps further to recognize the referent from other objects mentioned in the language input.

The task of grounding referring expression is to localize the region in the image given a referring expression. To solve this problem, joint embedding model is widely used in recent works~\cite{wang2016learning,rohrbach2016grounding,liu2017referring}. They model the conditional probability $P(o|r)$, where $r$ is the referent and $o$ is the appropriate visual object. Instead of modeling $P(o|r)$ directly, others~\cite{mao2016generation, yu2016modeling,nagaraja2016modeling,hu2016natural,luo2017comprehension,deng2018visual,yu2018rethinking} compute $P(r|o)$ by using the CNN-LSTM structure for language generation. The visual region $o$ maximizing $P(r|o)$ is considered to be the target region. Taking advantages of both the above mentioned approaches, Yu~\etal~\cite{yu2017joint} consider the joint-embedding model as a listener, CNN-LSTM as a speaker, and combine them to form a joint speaker-listener-reinforcer model to achieve state-of-the-art results. Instead of using holistic language feature  to do referring expression grounding, some recent work decompose the language input into different parts. Modular Attention Network (MAttNet) \cite{yu2018mattnet} decomposes expressions into three modules related to subject appearance, location, and relationship to other objects, rather than treating them as a single unit. This model then calculates an overall score dynamically from all the three modules with weights learned from the language based attention. Visual attention has been used to facilitate the subject and relationship modules to focus on relevant image regions. Compositional Modular Network (CMN)~\cite{hu2017modeling} is a modular deep architecture which divides the input language into vector representations of subject, relationship, and object with attention and then integrates the scores of these three modules into the final score indicating which region is more qualified for the given language input. Separating an entire sentence into several components and  analyzing these components using specific models makes the analysis more fine-grained. 

However, it is worth noting that natural language has a latent hierarchical structure. Facilitating such latent structure information would make the grounding model more reasonable and explainable. Our model steps further in this direction by taking the latent hierarchical structure of the language into account. We automatically compose a binary tree structure to parse the language and then perform visual reasoning along the tree in a bottom-up fashion by accumulating grounding confidence scores.

\subsection{Learning Tree Structures for Language}
In their NLP community, learning tree structures for sentences is becoming more and more popular in recent years. Bowman~\etal~\cite{bowman2016fast} built trees and compose semantics via a generic shift-reduce parser, whose training relies on ground-truth parsing trees. TreeRNNs combined with latent tree learning has been deemed as an effective approach for sentence embedding as it jointly optimizes the sentence embedding and a task-specific objective. For instance, Yogatama~\etal~\cite{yogatama2016learning} used REINFORCE algorithms~\cite{williams1992simple} to train the shift-reduce parser without ground truth. Instead of the shift-reduce parsers, Maillard~\etal~\cite{maillard2017jointly} used a chart parser, which is fully differentiable by introducing a softmax annealing but suffers from $\mathcal{O}(n^3)$ time- and space-complexity. Gumbel Tree-LSTM is a parsing strategy proposed by~\cite{choi2018learning}, which introduces Tree-LSTM and calculates the merging score for each adjacent node pair based on a learnable query vector and greedily merges the best pair with the highest score in the next layer. They introduced Straight-Through Gumbel-Softmax estimator~\cite{gumbelsoftmax} to soften a hard categorical one-hot distribution into a soft distribution so as to enable end-to-end training. 
Comparison between above mentioned models on several datasets, which is done by~\cite{williams2017learning}, shows that Gumbel Tree-LSTM achieves the best performance.
Our model facilitates the approach to learn the latent tree structure from a flat language input to achieve visual reasoning for the natural language grounding task. 

Tree structures for language have also been studied in the field of vision-language tasks. Xiao~\etal~\cite{xiao2017weakly} introduced the dependency parsing tree as a structural loss in visual grounding, thus the grounding results are expected to be more faithful to the sentence. Our work is fundamentally different from theirs as we explicitly perform grounding score calculation along the tree. In addition, note that our tree is more similar to constituency tree but not dependency tree. To minimize the biases in existing VQA datasets, Johnson~\etal~\cite{johnson2017clevr} proposed a diagnostic dataset CLEVR that tests a range of visual reasoning abilities. Questions in the dataset CLEVR are built using several categorical functions (\eg, Filter, Equal and Relate) by composing these simple building blocks. Johnson \etal~\cite{johnson2017inferring}
proposed a method which contains two main modules: program generator and execution engine. The program generator takes a sequence of words as inputs and outputs a program as a sequence of functions. The resulting sequence of the functions is then converted to a syntax tree for the execution of visual reasoning by making use of the fact that the arguments of each function are known. Hu~\etal~\cite{hu2017learning} proposed an End-to-End Module Networks (N2NMNs) containing two components: a layout policy, which inputs deep representation of a question and outputs both a sequence of structural actions and a sequence of attentive actions, and a network builder, which takes these two sequences as input and outputs an appropriately structured network to complete visual reasoning. All the above mentioned methods for VQA task seek to explore the latent structure of the input question.

\section{\textsc{RvG-Tree} Model}
We first define the problem of natural language grounding formally, and then introduce the \textsc{RvG-Tree} grounding model as illustrated in Fig.~\ref{fig:2} for a walk-through example. Finally, we show how to train \textsc{RvG-Tree} as an end-to-end neural network.

\subsection{Problem Definition}
We represent an image as a set of Region of Interest (ROI) features $\mathcal{I}=\{\mathbf{x}_1, \mathbf{x}_2, ..., \mathbf{x}_n\}$, where $\mathbf{x}_i\in\mathbb{R}^d$ is a $d$-dimensional feature vector, \eg, extracted from any deep vision model such as Faster R-CNN~\cite{ren2015faster}. Each ROI is a visual object detected in the image. We represent a natural language sentence as an $m$-length sequence $\mathcal{L} = \{\mathbf{w}_1, \mathbf{w}_2, ..., \mathbf{w}_m\}$, where $\mathbf{w}_i\in\mathbb{R}^b$ is a $b$-dimensional trainable word embedding vector, \eg, initialized from any word-vector models such as GloVe~\cite{pennington2014glove}. The task of grounding language $\mathcal{L}$ in image $\mathcal{I}$ can be represented as the following ranking problem:
\begin{equation}\label{eq:1}
\mathbf{x}^* = \argmax_i S(\mathbf{x}_i,\mathcal{L}),
\end{equation}
where $S(\cdot)$ is a grounding score function that evaluates the association between region $\mathbf{x}_i\in\mathcal{I}$ and language $\mathcal{L}$.

Designing a good score function for Eq.~\eqref{eq:1} is not trivial because it is challenging to exploit the compositional nature of the language: parsing the sentence into semantic structures that capture the implied referent (\ie, the target region) and the context (\ie, regions that help to distinguish the referent from others). Therefore, previous grounding models that only uses holistic sentence-level~\cite{mao2016generation} or phrase-level~\cite{plummer2017phrase} language features are straightforward but suboptimal. Recently, the triplet composition~\cite{hu2017modeling} is proposed to decompose the grounding score in Eq~\eqref{eq:1} into three sub-scores: referent (or subject), context (or object), and their pairwise relationship scores:
\begin{equation}\label{eq:2}
S(\mathbf{x}_i, \mathcal{L}) := S_s(\mathbf{x}_i, \mathbf{y}_s) + S_v(\mathbf{x}_v, \mathbf{y}_v) + S_p([\mathbf{x}_i,\mathbf{x}_v],\mathbf{y}_p),
\end{equation}
where $\mathbf{y}_s$, $\mathbf{y}_v$, and $\mathbf{y}_p$ are the $b$-dimensional language features (the same dimension as the word embedding) for the 3 linguistic roles: referent, context, and relationship, respectively. They are computed by soft-attention weighted sum over the word vectors in the sentence, where the attention weights are word-relevance to each of the linguistic roles. $\mathbf{x}_v\in\mathbb{R}^d$ is the ROI feature for the context. As illustrated in Fig.~\ref{fig:1a}, take ``a black dog on the left of the tree'' as an example with perfect language parsing and visual detection, $\mathbf{y}_s = 0.5\mathbf{w}_\textrm{black}+0.5\mathbf{w}_\textrm{dog}$, $\mathbf{y}_v = \mathbf{w}_\textrm{tree}$, and $\mathbf{y}_p = \mathbf{w}_\textrm{left}$, and $\mathbf{x}_v$ should be the ROI of ``tree''. Therefore, any region $\mathbf{x}_i$ of ``dog'' is expected to receive a higher score compared to the regions of other objects.

However, it is still not easy to obtain accurate $\mathbf{y}_s$, $\mathbf{y}_v$, $\mathbf{y}_p$, and $\mathbf{x}_v$ in Eq.~\eqref{eq:2}, especially, when the language consists of more complex compositions such as ``a black and white cat on top of the tree which is in front of a truck''. The reasons are due to the error-prone modules as follows.

\textbf{Language Composition}:
The \emph{referent}, \emph{context}, and \emph{relationship} compositions produced by off-the-shelf syntactic parsers do not always correspond to intuitive reasoning of visual grounding. For example, one of the objects in ``a black and white cat on top of the tree which is in front of a truck'' will be parsed, if perfectly, as ``the tree which is in front of a truck'', which is linguistically correct but visually difficult to learn the visual-semantic correspondence between a region and such a complex sub-expression. Therefore, we should further parse it into more fine-grained components for the ease of visual grounding.

\textbf{Context Localization}:
Due to the prohibitively high cost of annotating both referent and context in images~\cite{zhang2018grounding}, we have to guess the context in a weakly-supervised way, that is, during training, the context object is not localized as the referent with ground-truth bounding boxes. Moreover, the context is not a single region but a multinomial combination of all the possible regions mentioned in the language. For example, how to compose a comprehensive representation $\mathbf{x}_v$ for ``black and white'' and ``on top of the tree which is in front of a truck'' is still far from solved.

\subsection{\textsc{RvG-Tree} Construction}\label{sec:treeconstruction}
To address the two challenges introduced above, we propose to further decompose the grounding score in a recursive way by using a binary tree, allowing much more fine-grained visual reasoning. The motivations are two-fold: 1) the natural language can be generally divided into recursive components --- we can always use attributive clause to modify a noun when necessary, and each clause can be recursively parsed into two linguistic components, such as the (subject, object), (attribute, subject), or (preposition, subject) pairs.  2) by using trees, we can do more fine-grained localization with simpler expressions and thus simple grounding scores can be accumulated along the tree in a bottom-up fashion.

\begin{figure}[t]
\begin{subfigure}{.25\textwidth}
  \centering
  \includegraphics[width=.8\linewidth]{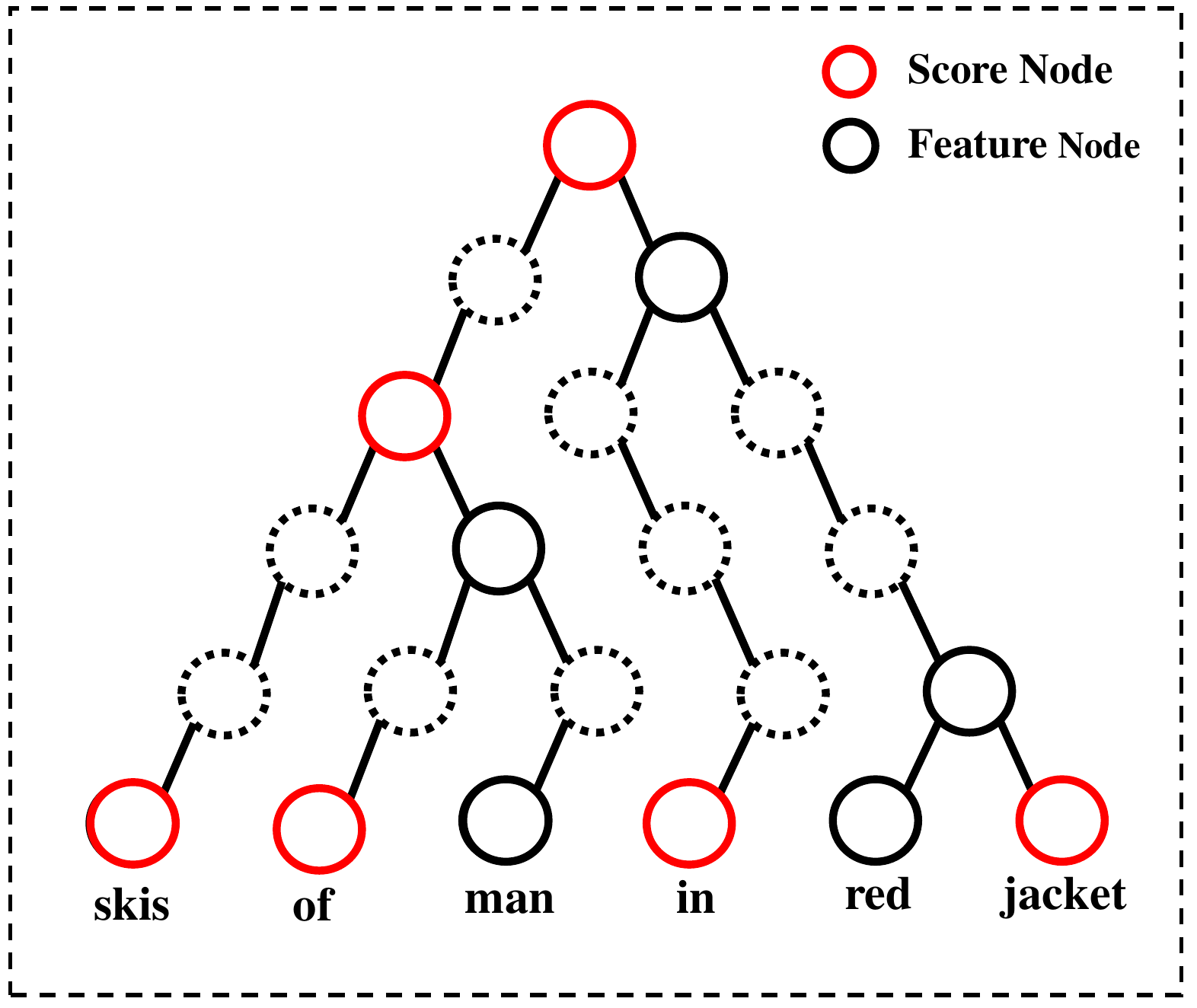}
  \caption{Tree Construction}
  \label{fig:3a}
\end{subfigure}%
\begin{subfigure}{.25\textwidth}
  \centering
  \includegraphics[width=.8\linewidth]{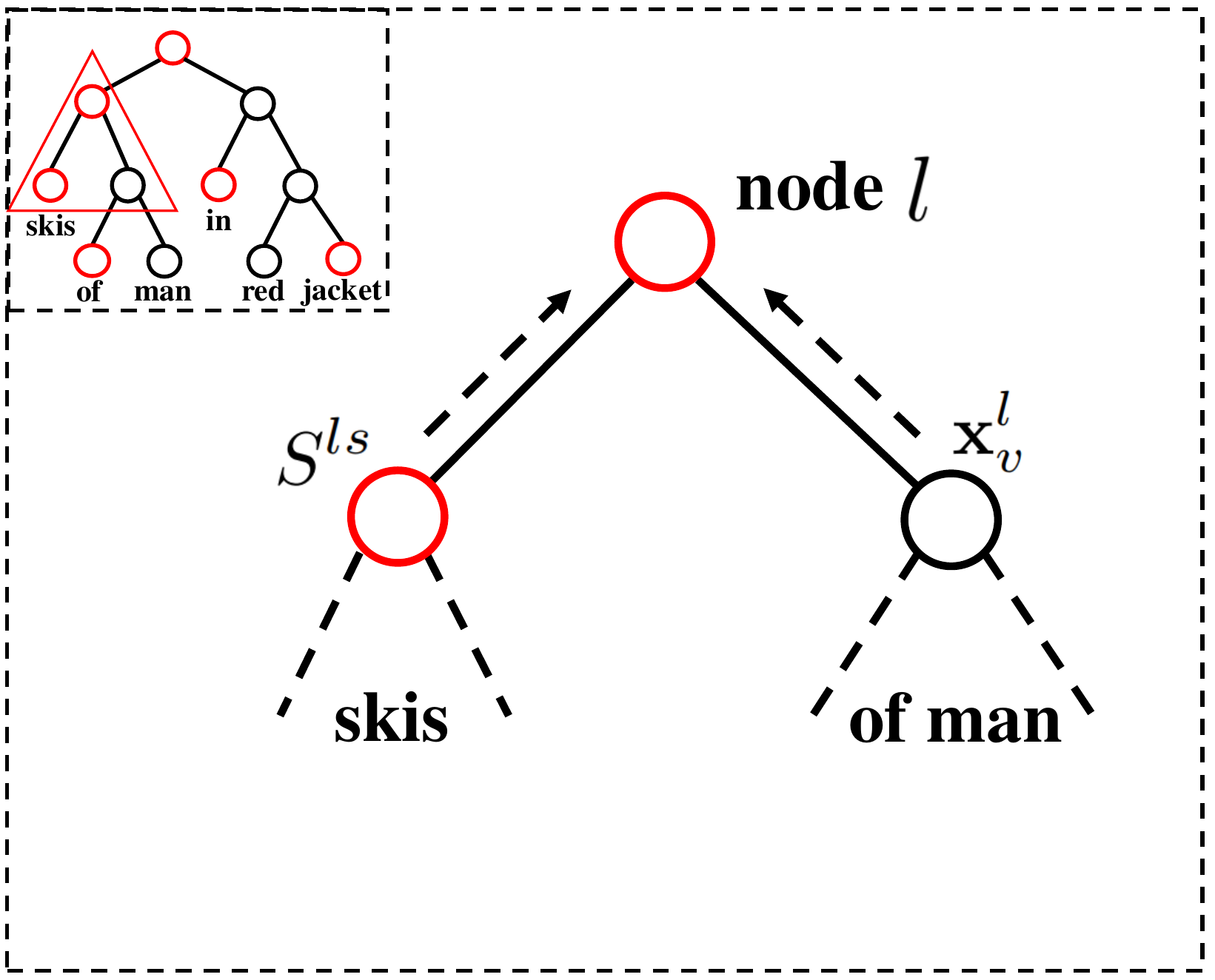}
  \caption{Recursive Unit}
  \label{fig:3b}
\end{subfigure}
\vspace{-2mm}
\caption{ (a). Given a flat sentence, RvG-Tree computes a score for every parent candidate indicating how qualified each candidate is to be merged in the next layer. By operating selection and merging recursively until RvG-Tree reaches the root node, the tree is constructed and will be used in the inference part. \quad (b). RvG-Tree calculates grounding confidence scores in a recursive way. Taking the $\mathit{l}$-th node as an example, according to Eq.~\eqref{eq:3}, the grounding score returned by it is a sum of the scores from itself and its ``score child''. }
\vspace{-4mm}
\label{fig:tree}
\end{figure}

Before we construct the tree, we prune the sentence by discarding some determiners and symbols such as ``a, an, another, any, both, each, either, those, that''. We find that this pruning does not affect the overall performance while boost the speed. Similar to the method in~\cite{choi2018learning}, RvG-Tree calculates a validity score indicating how valid a composition is for every parent candidate. Composition here means to merge two adjacent nodes into a parent. Based on the validity score, the model recursively selects compositions in a bottom-up fashion, until it reaches the root. Fig.~\ref{fig:3a} is a walk-through example of \textsc{RvG-Tree} construction for the sentence ``skis of man in red jacket'' in Fig.~\ref{fig:2}. The first merging happens at ``red'' and ``jacket'', then the merged node together with other nodes: ``skis'', ``of'', ``man'', and ``in'', are the input for the next merging process, and ``of'' and ``man'' are merged. We repeat this process until we have two nodes left: one merged from ``skis of man'' and the other one merged from ``in red jacket''.

Formally, we first need to embed each node into features and then use them to decide which two nodes to merge in a computational way. We start from the leaf nodes, where each one is represented as the word embedding $\mathbf{y}_l$ in the sentence $\mathcal{L}$.  To encode the contextual information of the words in sentence, we use a bi-directional LSTM (BiLSTM)~\cite{schuster1997bidirectional} to obtain the initial $l$-th node feature $\mathbf{v}^{t=1}_l$ as:
\begin{equation}
\mathbf{v}^1_l = \begin{bmatrix}
\mathbf{h}^1_l\\ \mathbf{c}^1_l \end{bmatrix}
= \textrm{BiLSTM}(\mathbf{y}_l, \mathbf{h}^1_{l-1},  \mathbf{h}^1_{l+1}),
\label{eq:leafbilstm}
\end{equation}
where $\mathbf{h}^1_l$ and $\mathbf{c}^1_l$ are the hidden and memory cell vectors of the BiLSTM. Then, we can use $\mathbf{v}^1_l$ to merge two of them for the next layer $t=2$ by using Eq.~\eqref{eq:parentscore}, which will be discussed soon. Next, we introduce how to obtain the node features for layers $t\leq 2$.  

Without loss of generality, as shown in Fig.~\ref{fig:3a}, suppose ``red'' and ``jacket'' are merged as a new node for the next layer $t=2$, then all the other leaf nodes are upgraded to $t=2$ for the next merging step (the dashed nodes in Fig.~\ref{fig:3a}):
\begin{equation}
\mathbf{v}^{t+1}_j =  
\begin{cases}
\mathbf{v}^{t}_j & j < l\cr
\begin{bmatrix}
\mathbf{h}^{t+1}; \mathbf{c}^{t+1} \end{bmatrix} = \textrm{TreeLSTM}(\mathbf{v}^{t}_j, \mathbf{v}^{t}_{j+1} ) & j = l\cr
\mathbf{v}^{t}_{j+1} & j > l
\end{cases}
\label{eq:update}
\end{equation}
where $\mathbf{h}^t$ and $\mathbf{c}^t$ are the hidden and memory cell vectors from the Tree LSTM network (TreeLSTM)~\cite{socher2013recursive}, which is a simple extension of the original LSTM by concatenating the children hidden states as the input hidden states.

Now, we introduce how to merge any two adjacent nodes. We introduce a trainable parameter $\mathbf{\theta}_p$ to measure the validity of a parent. Specifically, we use $\mathbf{\theta}^T_p \mathbf{v^t_l}$ as the unnormalized validity score of a candidate parent representation in Eq.~\eqref{eq:update}. Then, we decide whether to merge its two candidate children by selecting the largest (\ie, argmax) softmax normalized score:
\begin{equation}
s_l = \frac{\exp(\mathbf{\theta}^T_p {\mathbf{v}}^{t}_l) }
{\sum_j \exp(\mathbf{\theta}^T_p {\mathbf{v}}^t_j)}.
\label{eq:parentscore}
\end{equation}
We repeat this procedure until we reach the root node of the tree, \ie,  the final \textsc{RvG-Tree} structure. 

Note that the node feature embedding functions described in Eq.~\eqref{eq:leafbilstm} and Eq.~\eqref{eq:update} are differentiable, but the merge procedure by using Eq.~\eqref{eq:parentscore} is not, due to the greedy argmax. To tackle the discrete nature of the tree structure construction, we deploy the Gumbel-Softmax trick detailed in Section~\ref{sec:stgs}.

\subsection{Recursive Grounding}\label{sec:recursivegrounding}
Given the constructed \textsc{RvG-Tree} described in the previous section, we can accumulate the grounding confidence scores according to the language composition along the tree in a bottom-up fashion. Without loss of generality, suppose we are interested in calculating the $l$-th node (Fig.~\ref{fig:3b}), which has two children nodes: \emph{score} node $ls$ and \emph{feature} node $lv$ (cf. Section~\ref{sec:sfnode}). Then, the grounding score returned by the $l$-th node is defined in a recursive fashion:
\begin{equation}\label{eq:3}
\begin{split}
&S^l(\mathbf{x}_i, \mathcal{L}_l) := \overbrace{S^l_s(\mathbf{x}_i, \mathbf{y}^l_s)  + S^l_p([\mathbf{x}_i,\!\!\!\!\!\!\!\!\!\!\!\!\!\!\!\!\!\!\!\!\!\!\underbrace{\mathbf{x}_{lv}}_{\text{feature returned from node $lv$}}\!\!\!\!\!\!\!\!\!\!\!\!\!\!\!\!\!\!\!\!\!],\mathbf{y}^l_p)}^{\text{score calculated at node $l$}}\\
&+\!\!\!\!\!\!\!\underbrace{S^{ls}(\mathbf{x}_i,\mathcal{L}_{ls})}_{\text{score returned from node $ls$}}\!\!\!\!\!\!.
\end{split}
\end{equation}

From the ``divide \& conquer'' perspective, the ``dirty'' job (\ie, conquer) is done by the ``score calculated at node $l$'' terms, and the ``easy'' ones (\ie, divide) are just to ask the $lv$ and $ls$ children to give us ``feature returned from node $lv$'' and  ``score returned from node $ls$'', and thus the overall reasoning can be performed in a bottom-up fashion. Interestingly, Eq.~\eqref{eq:3} can be viewed as a more generic and hierarchical formulation of the widely-used triplet composition~\cite{hu2017modeling,yu2016modeling,yu2018mattnet} as in Eq.~\eqref{eq:2}; however, the key difference is that our composition is achieved via an explicitly recursive tree, while the previous one is learned in an implicitly flat fashion.

\noindent\textbf{Complexity}. Compared to holistic or simple compositional methods such as Hu~\textit{et al.}~\cite{hu2017modeling} and Yu~\textit{et al.}~\cite{yu2018mattnet}, the proposed recursive grounding in Eq.~\eqref{eq:3} is more computationally expensive but the overhead is linear to sentence length and thus affordable. Suppose their computational cost is unit 1 and the sentence length is $N$, the number of score calculation is the number tree nodes: $\mathcal{O}(2N)$.

Next, we will discuss the design and notation details of Eq.~\eqref{eq:3} as follows.

\subsubsection{Score Node \& Feature Node Definition}\label{sec:sfnode}
According to the primal score of Eq.~\eqref{eq:1}, a grounding score is to measure the association between any region $\mathbf{x}_i$ and a language sentence $\mathcal{L}$. However, as our recursive grounding will go through every word or sub-sequence in the sentence, it is not always reasonable to accumulate every score. To this end, we introduce the score and feature nodes that are specially designed for recursive grounding, as illustrated in Fig.~\ref{fig:3b}.

\noindent\textbf{Score Node}. If a node $ls$ is a score node, its score  will be accumulated to the higher layer. In particular, every root is a score node. Compared to the following introduced feature node, a score node calculates the grounding score $S^{ls}$ as in Eq.~\eqref{eq:3} and deliver to the higher-layer node. Thanks to the score nodes, we can relax the unreasonable cases in visual grounding that every language component should correspond to a visual region. We will further discuss this intuition later in Section~\ref{sec:3.3.2}.

\noindent\textbf{Feature Node}. If a node $lv$ is a feature node, it first calculates the grounding score $S^{lv}$ as in Eq.~\eqref{eq:3}, and then aggregates a weighted sum over the region features, where the weights are normalized by the grounding score:
\begin{equation}\label{eq:visualctx}
\begin{split}
\mathbf{x} &= \sum_i \frac{\exp\left(S^{l}(\mathbf{x}_i,\mathcal{L}_{l})\right)}{\sum_j\exp\left(S^{l}(\mathbf{x}_j,\mathcal{L}_{l})\right)} \mathbf{x}_i,\\
\mathbf{x}_{lv}&\leftarrow \mathbf{x}.
\end{split}
\end{equation}
Note that the above feature assignment indicates that the weighted feature $\mathbf{x}$ at the current node $l$ is delivered to its parent node $l$ and considered as the output of the feature node $\mathbf{x}_{lv}$ in Eq.~\eqref{eq:3}. In the view of the hierarchical feature representations in deep networks, we should feed-forward the visual features in a bottom-up fashion, \ie, the feature node. On the other hand, the score node is analogous to the loss that accumulated from intermediate features~\cite{szegedy2015going}.

\subsubsection{Score Node \& Feature Node Classification}\label{sec:3.3.2}
To determine whether child node $1$ (or $2$) is the feature node $lv$ and the other one is the score node $ls$, we use the following binary softmax to be the ``feature node'' probability $p_{v1}$:
\begin{equation} 
p_{v1} =  \frac{\exp(\theta^T_{v} \mathbf{v}_1)}
{\exp(\theta^T_{v} \mathbf{v}_1) + \exp(\theta^T_{v}  \mathbf{v}_2) }
\label{eq:rc},
\end{equation}
where $\theta_{v}$ is the trainable parameter, $\mathbf{v}_{1/2}$ is the children node feature exactly as the same as in Eq.~\eqref{eq:update}. Note that we also have the following probabilities:
\begin{equation}
    p_{v2} = p_{s1} = 1-p_{v1},~~~p_{s2} = p_{v1}, 
\end{equation}
where $p_{s\cdot}$ is probability of score node. Similar to the discrete policy that causes non-differentiability in tree construction as in Eq.~\eqref{eq:parentscore}, we deploy Gumbel-Softmax to resolve this issue raised by Eq.~\eqref{eq:rc}.

\begin{figure}[t]
\begin{subfigure}{.25\textwidth}
  \centering
  \includegraphics[width=.9\linewidth]{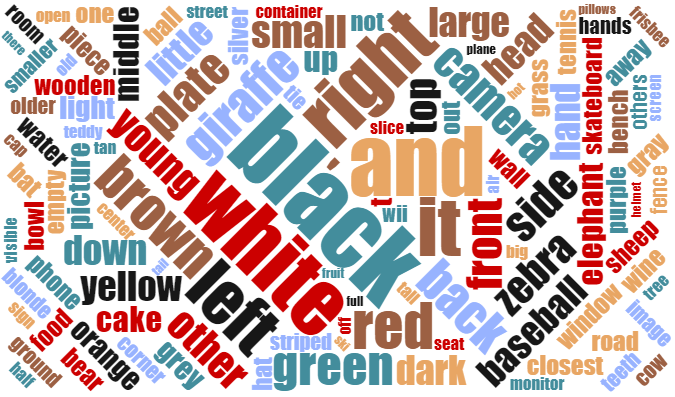}
  \caption{Feature Node}
  \label{fig:score_cloud}
\end{subfigure}%
\begin{subfigure}{.25\textwidth}
  \centering
  \includegraphics[width=.9\linewidth]{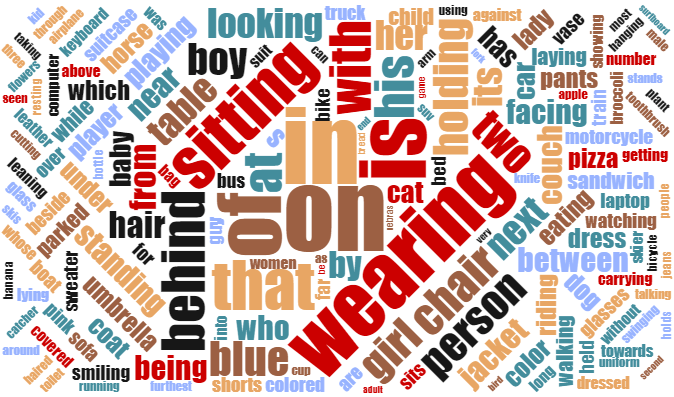}
  \caption{Score Node}
  \label{fig:feature_cloud}
\end{subfigure}
\caption{Word cloud visualizations of what nodes are likely to be classified as (a) feature node or (b) score node. Every word frequency is increased by 1 if it is a leaf under the feature or score node. We can see that feature nodes are generally related to visual concepts while score nodes are generally not visual.}
\label{fig:cloud}
\end{figure}

Fig.~\ref{fig:cloud} illustrates what nodes are likely to be classified as the feature or score nodes. We can see that nodes which have more visual leaf words like adjectives such as colors are more likely to be the feature nodes, and those which have more non-visual leaf words like relationships (``behind'' and ``sitting'') are more likely to be the score nodes.
 
\subsubsection{Language Feature}
We use language feature to localize the corresponding visual regions referred in the language. Essentially, we would like to calculate the multimodal association between the visual features and the language features. We denote $\mathbf{y}^l_s$ as the language feature used to associate with a single region feature (\ie, $\mathbf{x}_i$) and $\mathbf{y}^l_p$ to associate with a pairwise visual feature (\ie, $[\mathbf{x}_i,\mathbf{x}_{lv}]$). Specifically, the language feature is calculated as a soft-weighted sum of the corresponding word embeddings in the sub-sequence $\mathcal{L}_l$:
\begin{equation}\label{eq:12}
\mathbf{y}^l_s = \sum\limits^{|\mathcal{L}_l|}_{i=1} \alpha_{si} \mathbf{w}_i,~~\mathbf{y}^l_p = \sum\limits^{|\mathcal{L}_l|}_{i=1} \alpha_{pi} \mathbf{w}_i,
\end{equation}
where $\mathbf{w}_i$ is the $i$-th word embedding vector and $\alpha$ is the word-level attention weights:
\begin{equation}\label{eq:13}
\alpha_{si} = \frac{\exp\left(\theta^T_s\mathbf{v}_i\right)}{\sum_j\exp\left(\theta^T_s\mathbf{v}_j\right)}, ~~~\alpha_{pi} = \frac{\exp\left(\theta^T_p\mathbf{v}_i\right)}{\sum_j\exp\left(\theta^T_p\mathbf{v}_j\right)},
\end{equation}
where $\theta$ is the trainable parameter and $\mathbf{v}$ is the leaf node feature that is the same as in Eq.~\eqref{eq:update}\&\eqref{eq:rc}. The reason why we need word-level attention for extracting the language feature is because not all the words are related to the visual feature. Hence, suppressing irrelevant words will help the multimodal association. It is worth noting that the reason why we use the sum of word-level embeddings while not the BiTreeLSTM hidden vectors is because the latter is too diverse in the limited sentence pattern in our training scenario; however, the former is more stable as the diversity of words is significantly smaller than that of word compositions in sentence.  

\subsubsection{Score Functions}
The score function $S^l_s$ indicates how likely $\mathbf{x}_i$ is the referent given $\mathbf{y}^l_s$ and $S^l_p$ shows how the pair-wise visual feature $[\mathbf{x}_i, \mathbf{x}_{lv}]$ matches the relationship described in $\mathbf{y}^l_p$. Formally, they are defined as the following simple two-layer MLPs~\cite{hu2017modeling,zhang2018grounding}:
\begin{equation}
S^l_s(\mathbf{x}_i, \mathbf{y}^l_s) = \theta^T_{s1}\left[ \mathrm{L2Norm}\left(\theta^T_{s2}\mathbf{x}_i \odot \mathbf{y}^l_s\right)\right]
\label{eq:score_s},
\end{equation}

\begin{equation}
S^l_p([\mathbf{x}_i,\mathbf{x}_{lv}], \mathbf{y}^l_p) = \theta^T_{p1}\left[ \mathrm{L2Norm}\left(\theta^T_{p2}[\mathbf{x}_i,\mathbf{x}_{lv}] \odot \mathbf{y}^l_p\right)\right]
\label{eq:score_p}
\end{equation}
where the $\theta$s are the trainable parameters, $\odot$ is element-wise multiplication, L2Norm is used to normalize the vector as unit L2 norm. 

\subsubsection{Leaf Case}
The recursive Eq.~\eqref{eq:3} will arrive at the one layer above the leaves, that is,  when the two children nodes are words, we may encounter extreme cases that the words cannot be visually grounded such as ``with'', ``of'', and ``is'', causing difficulties in interpreting the scores $S^l_s$ and $S^l_p$. Fortunately, in these cases, the score functions defined in Eq.~\eqref{eq:score_s} and ~\eqref{eq:score_p} will calculate similar scores for each region, as none of them is visually related to the words. As a result, accumulating such trivial scores would not affect the overall score ranking. 

When the recursion in Eq.~\eqref{eq:3} arrives at the leaves, $S^{ls}$ will calculate a grounding score for an empty sentence. Thus, we define the exit of the recursion as:
\begin{equation}
S^{ls}(\mathbf{x}_i,\mathcal{L}_{ls}) = S^{ls}(\mathbf{x}_i,\phi) = 0, ~~ \textrm{if}~~l=~\textrm{leaf}.
\label{eq:score_leaf}
\end{equation}

\begin{algorithm}
\SetAlgoLined
\SetKwInOut{Input}{Input}
\SetKwInOut{Output}{Output}

\Input{Language sentence, Image features}
 Prune the language sentence\;
 Embed words into node features as Eq.~\eqref{eq:leafbilstm}\;
 Initialize grounding scores as Eq.~\eqref{eq:score_leaf}\;
 \For{\rm{Each layer}}{
  Find which two adjacent nodes to merge as Eq.~\eqref{eq:parentscore}\;
  Classify score node and feature node as Eq.\eqref{eq:rc}\;
  Update node features as Eq.~\eqref{eq:update}\;
  Update grounding scores for each node as Eq.~\eqref{eq:3}\;
}
\Output{Grounding score of root}
 \caption{\textsc{RvG-Tree Grounding Pipeline}}
\label{alg}
\end{algorithm}

\subsection{\textsc{RvG-Tree} Training}
The inference of the \textsc{RvG-Tree} model is summarized in Algorithm~\ref{alg}. The model can be trained end-to-end in a supervised fashion when the ground truth region $\mathbf{x}_{g}$ referred in the language $\mathcal{L}$ is given. To distinguish the referring region from others, the model is expected to output a high $S(\mathbf{x}_g, \mathcal{L})$ for the ground-truth region and a low $S(\mathbf{x}_i, \mathcal{L})$ whenever $i \neq g$. Therefore, we train our model using the following cross-entropy loss:
\begin{equation}
L(\Theta) = - \log \frac{\exp\left(S(\mathbf{x}_{g},\mathcal{L})\right)}{\sum_i \exp(S\left(\mathbf{x}_{i},\mathcal{L})\right)}
\label{eq:loss},
\end{equation}
where $\Theta$ denotes all the trainable parameters in our model. Note that this loss is also called the Maximum Mutual Information (MMI) training in the pioneering work~\cite{mao2016generation}, as it is the same as maximizing the mutual information between the referent region and others (with the assumption of a uniform prior). The purpose is to ground the referent unambiguously by penalizing the model if it grounds other regions with high scores. Within a similar spirit, we can reformulate Eq.~\eqref{eq:loss} into a large-margin triplet loss as in~\cite{yu2017joint} with hard-negative sample mining. However, in our experiments, we observed only marginal performance gain but more tricky learning rate adjustment. Thus, we use Eq.~\eqref{eq:loss} as the overall training objective in this paper. 

\subsubsection{Straight-Through Gumbel-Softmax Estimator}\label{sec:stgs}
Note that it is prohibitive to use stochastic gradient descent (SGD) that back-propagates gradients of Eq.~\eqref{eq:loss} to update the parameters of our model. The reason is that the gradients are blocked in the steps that make discrete policies where we greedily choose a parent node according to Eq.~\eqref{eq:parentscore} and decide which child is the feature node according to Eq.~\eqref{eq:rc}. To bridge the gradients over the gap raised by the discrete policies, we deploy the Straight-Through (ST) Gumbel-Softmax estimator~\cite{gumbelsoftmax}, which takes different paths in the forward and backward propagation by replacing the discrete $\argmax$ function with a differentiable and re-parameterized softmax function. Formally, given unnormalized probabilities $\{p_1, \ldots, p_n\}$, a sample $b_i \in \{b_1, \ldots, b_n\}$ from the Gumbel-Softmax distribution is drawn by $b_i\sim \pi_i$:

\begin{equation}
\pi_i = \frac{\exp((\log(p_i) + g_i) / \tau)}{\sum^n_{j=1} \exp ((\log(p_j) + g_j) / \tau)},
\end{equation}
where $g_i = -\log(-log(u_i))$ and $u_i\sim \textrm{Uniform}(0,1)$. $g_i$ is called the Gumbel noise perturbing each $\log(p_i)$ and $\tau$ is a temperature parameter which diminishes to zero, a sample from the Gumbel-Softmax distribution becomes a \textit{cold} one resembling the one-hot sample.   

Then, the straight-through (ST) gradient estimator is used as follows: in the forward propagation, $b_i$ is sampled by argmax; in the backward propagation, $b_i$ has a continuous value:
\begin{equation}\label{eq:case}
b_i = \begin{cases}
\argmax_{j} b_j, & \textrm{forward prop}\cr
 \pi_i, &  \textrm{backward prop}
\end{cases}.
\end{equation}
Intuitively, the estimator applies some random explorations (controlled by $u_i$) to select the best policy greedily in the forward pass, and it back-propagates the errors to all policies with a scaling factor $\pi_i$. Note that the noise in forward propagation of Eq.~\eqref{eq:case} is turned off in the test phase. Though this ST estimator is biased, it is shown to perform well in previous work~\cite{chung2016hierarchical} and our experiments.  Gumbel-Softmax and its ST estimator is a re-parameterization trick for feature-based random variable where the output of a random choice is a feature while not a discrete layout. Thanks to this trick, the ST estimator can be considered as a soft-attention mechanism that efficiently back-propagates errors for all possible discrete tree structures, smartly avoiding from sampling the prohibitively large layout space such as REINFORCE~\cite{williams1992simple}. In our experiments, we found that REINFORCE with Monte-Carlo sampling does not converge with even very small learning rate such as 1e-6.

\subsubsection{Supervised Pre-Training}
Minimizing the loss function in Eq.~\eqref{eq:loss} from scratch is challenging: one need to simultaneously learn the all the parameters, especially those for the tree construction and feature/score node selection policies, which may suffer from a weaker stability and be trapped in a local optimum, greatly affecting the performance of our \textsc{RvG-Tree}. Therefore, we would like to apply a common practice: we first use the supervised pre-training to find a fair solution, \ie, a good exploitation, and then use the end-to-end straight-through Gumbel Softmax as weakly supervised fine-tuning to achieve better exploration.  However, there is no such an expert policy to generate a \textsc{RvG-Tree} like binary tree. To tackle this challenge, we borrow a third-part toolkit: Stanford CoreNLP (SCNLP)~\cite{stanfordcorenlp}, which contains a constituency parser that takes a cleaned flat sentence as input and outputs a multi-branch constituency tree, where the children of a node are words constituting a phrase, \eg, ``furry black dog''. SCNLP cleans the input sentence using \textit{pos tag} before feeding it into the constituency parser by discarding punctuations and articles that bring unnecessary redundancy in the tree. 

\begin{figure}[t]
\begin{subfigure}{.24\textwidth}
  \centering
  \includegraphics[width=.89\linewidth]{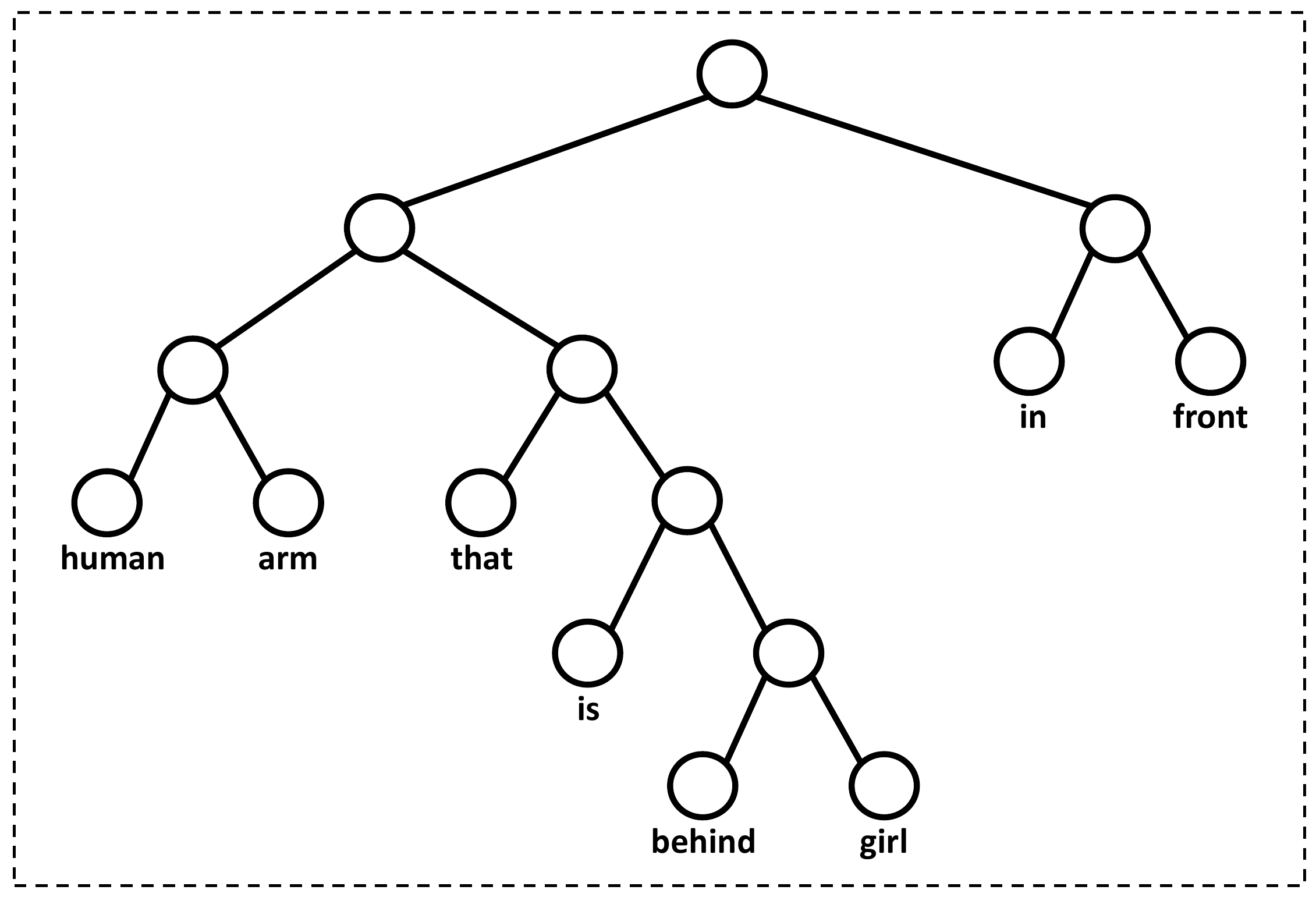}
  \caption{Expert Tree}
  \label{fig:original}
\end{subfigure}%
\begin{subfigure}{.24\textwidth}
  \centering
  \includegraphics[width=.89\linewidth]{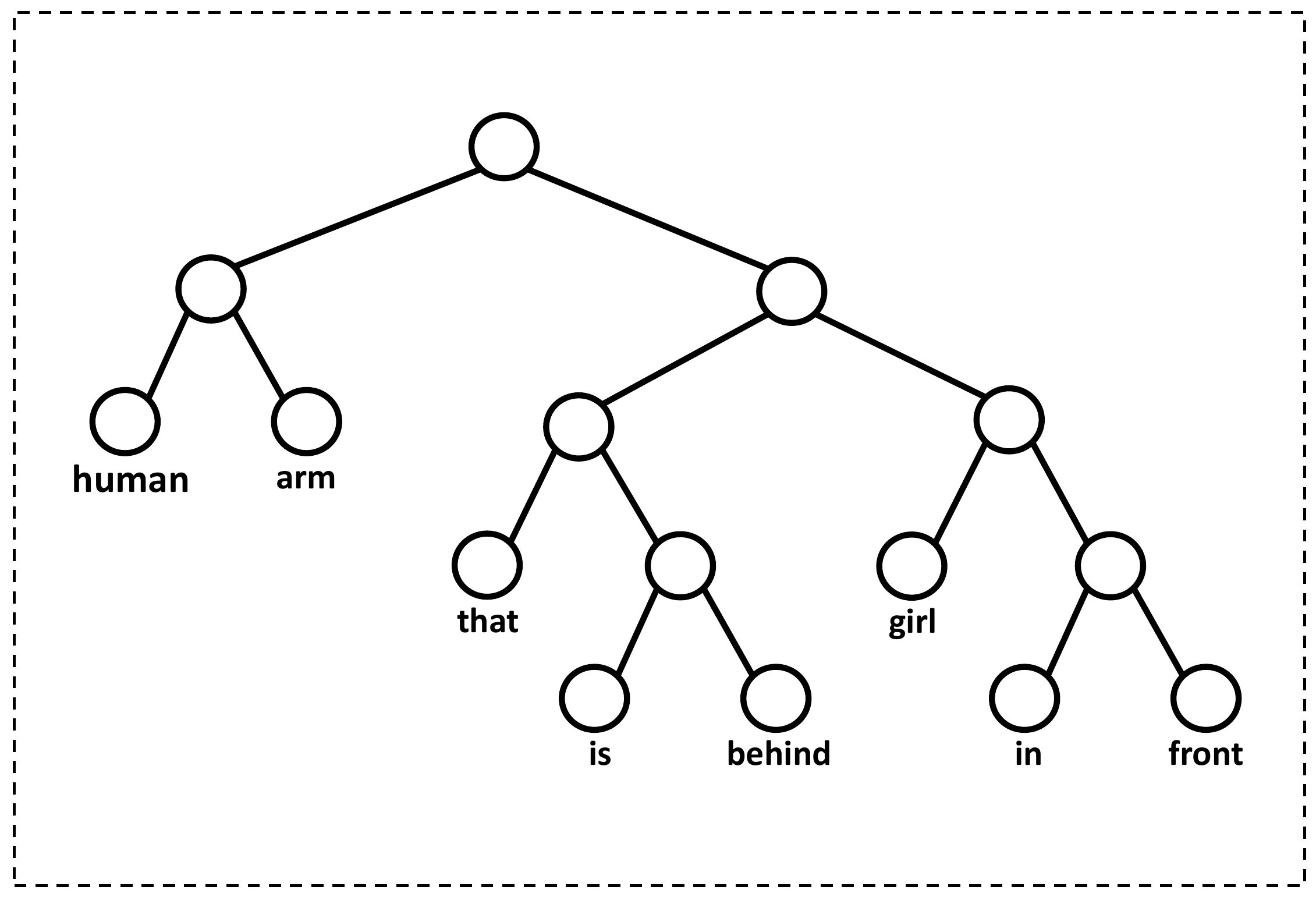}
  \caption{Fine-tuned Tree}
  \label{fig:finetuned}
\end{subfigure}
\caption{ Examples of binary tree structures from (a) SCNLP constituency parser and (b) overall fined-tuned by Straight-Through Gumbel-Softmax estimator. We can see that the tree structure is  more meaningful after fine-tuning.  }
\label{fig:fine_tree}
\end{figure}

To transform the multi-branch constituency tree into a binary one, we apply a simple separation rule: for a sub-tree with $n$ children, from left to right, we group every two consecutive words that constitutes a sub-binary-tree, and the left one, if any, is upgraded to be a single sub-tree with itself as the root. For example, the five children ``a'', ``furry'',  ``and'',  ``black'', ``dog'' are separated into ``a furry'', ``and black'', and ``dog'', and are further merged to ``a furry and black'' and ``dog''. Thus, we use this binary tree as the expert layout to train Eq.~\eqref{eq:parentscore} with supervision.  Fig.~\ref{fig:fine_tree} illustrates the differences between an expert tree and a resultant tree after fine-tuning. First, the expert rules divides the sentence into two chuncks that are difficult for grounding: ``human arm that is behind girl'' and ``in front''; however, after fine-tuning, we can construct the tree in a more meaningful way: ``human arm'' that is the referent and ``that is behind girl in front'' as the context to be further parsed.

\section{Experiments}
We conducted extensive experiments on three benchmarks of referring expression comprehension, \ie, grounding the referent object described in the language. The motivation of our experimental design is to answer the following three questions:
\begin{itemize}[leftmargin=.1in]
\item
Is the tree structure better than holistic and triplet language models? 
\item
Is the recursive grounding score effective?
\item
Is \textsc{RvG-Tree} explainable?
\end{itemize}
\subsection{Datasets}
\textbf{RefCOCO}~\cite{yu2016modeling}. It contains 142,210 expressions, which are collected using an interactive game, for 50,000 object instances in 19,994 images. All expression-referent pairs in this dataset are split into four mutually-exclusive parts: train, validation, Test A and Test B. This dataset allots 120,624 and 10,834 pairs to the train and validation part, respectively. Test A is alloted with 5,657 images, where each image has multiple people and Test B contains the rest 5,095 expression-referent pairs, where each image contains multiple objects.

\noindent\textbf{RefCOCO+}~\cite{yu2016modeling}. It contains 141,564 referring expressions for 49,856 referents in 19,992 images. The referring expressions are collected in the same way as RefCOCO. It describes the referents with only appearance information by excluding absolute location words, making it different from RefCOCO. The train, validation, Test A, and Test B sections mutually exclusively contain 120,191, 10,758, 5726, and 4,889 expression-referent pairs, respectively.

\noindent\textbf{RefCOCOg}~\cite{mao2016generation}. It contains 95,010 expressions for 49,822 referents in 25,799 images. The expressions containing both appearance and location expressions, which are collected in a non-interactive way, are longer than those in RefCOCO and RefCOCO+. 
There are two kinds of data separations for RefCOCOg. The first one~\cite{mao2016generation} has no testing split released, so most recent works evaluate their models on the validation section. It is worth noting that the first partition randomly separates objects into training and validation sets, which leads to the fact that the images in both training and validation sets are not mutually exclusive. The second one~\cite{nagaraja2016modeling} randomly splits images into training, validation and testing sets. We report our experimental results on both of them. RefCOCOg has significantly richer language and hence is more challenging than the previous two datasets. 

\subsection{Settings and Metrics}
We set the length of each sentence in RefCOCO, RefCOCO+, and RefCOCOg, to 10, 10, 20, respectively, since such lengths can cover almost 95 percent sentences on all datasets. We use 'pad' symbol to pad expression sequences whose lengths are less than the set length. We use three specific vocabularies for the three datasets and the vocabulary sizes are 1,969, 2,596, and 3,314 for RefCOCO, RefCOCO+, and RefCOCOg, respectively. Word frequency in vocabularies was counted in all expressions and we discarded them that appeared less than 5 times in the entire dataset, then we replaced them with `unk' in the vocabulary. Note that these `unk's were still evaluated with the merge score in Eq.~\eqref{eq:parentscore} and involved in the Gumbel-Softmax in training and softmax in test with zero-out mask, that is, the `unk's were not considered in greedy merge. We used GloVe pre-trained word vectors~\cite{pennington2014glove} to initialize our word vectors. However, random initialized word vectors did not significantly degrade the performance.

We used ROI visual features annotated by MSCOCO for all three datasets. These ROIs are represented by 2048-d vectors, which are the fc7 output of a ResNet-101 based Faster-RCNN~\cite{ren2015faster} trained on MSCOCO, 1024-d vectors, which are the pool5 output of the same Faster-RCNN, and 5-d vectors, which indicate the location of ROI in an image. Note that the goal of our experiments is to diagnose the visual reasoning capability of the grounding model, therefore, we did not use the most recent strong visual attribute features as in~\cite{yu2018mattnet}. However, \textsc{RvG-Tree} is compatible to any visual feature input. 

We used Adam\cite{kingma2014adam} with initial learning rate 0.001, $\alpha=0.8 $, $\beta=0.999$, and $\epsilon=\mathrm{e}^{-8}$, as our optimizer. We set 128 images to mini-batch size. For each sentence grounding, we calculated the intersection-over-union (IoU) of the selected bounding box with the ground-truth bounding box and considered the one with IoU larger than 0.5 as correct. We compute the fraction of correctly grounded test expressions as the grounding accuracy (\ie, Top-1 Accuracy).

\subsection{Ablative Studies}
We conducted extensive ablative studies of \textsc{RvG-Tree} to justify our proposed design and training strategy. The ablations and their motivations are detailed as follows.
\begin{itemize}[leftmargin=.1in]
\item
\textbf{Chain}: We used BiLSTM to encode the sentence. Every word has two representations: 1) the 2048-d concatenation of its corresponding two-directional LSTM hidden vectors, and 2) the 300-d word embeddings. The first representation is used to calculate the word-level soft-attentions and then the language feature is represented as the soft-attention weighted average of the word embeddings. This ablation ignores the structure information of the language.
\item
\textbf{\textsc{RvG-Tree}-Fix}: We used TreeLSTM to encode the sentence. The binary tree is the constituent parsing tree result from Stanford Parser~\cite{zhu2013fast}. Similar to Chain, every word has 1) word embedding representations and 2) LSTM hidden state representations.
\item
\textbf{\textsc{RvG-Tree}-Scratch}: This is the full \textsc{RvG-Tree} model without the binary tree expert supervision, \ie, the tree is constructed from scratch. 
\item
\textbf{\textsc{RvG-Tree}/Node}: The tree is constructed in the same way as the full \textsc{RvG-Tree} model. The difference is that we ignore the first two scores in Eq.~\eqref{eq:3}. We used this ablation to justify that the in-node score is an essential complementary to the bottom-up score accumulation.

\begin{table*}[ht]
\footnotesize
\begin{center}
\begin{tabular}{| l | c | c | c || c | c | c || c | c |}
\hline
& \multicolumn{3}{c}{RefCOCO} & \multicolumn{3}{|c|}{RefCOCO+} & \multicolumn{2}{|c|}{RefCOCOg}\\
\cline{1-9}
& val & testA & testB & val & testA & testB & val & test\\
\hline\hline
Chain             & 80.14 & 80.54 & 79.61 & 65.77 & 66.80 & 60.97 & 72.77 & 71.69 \\
\textsc{RvG-Tree}-Fix        & 81.52 & 80.77 & 80.53 & 66.33 & 68.16 & 62.53 & 73.69 & 73.07 \\
\textsc{RvG-Tree}-Scratch & 79.65 & 79.22 & 79.33 & 65.01 & 66.12 & 61.12 & 71.83 & 72.00 \\
\textsc{RvG-Tree}/Node  & 82.93 & 82.41 & 82.30 & 67.76 & 69.33 & 64.47 & 74.10 & 74.36 \\
\textsc{RvG-Tree}/S     & 82.24 & 81.12 & 80.91 & 67.32 & 68.87 & 64.05 & 74.21 & 73.18 \\
\textsc{RvG-Tree}/F     & 82.50 & 81.79 & 81.49 & 67.48 & 69.37 & 64.29 & 74.12 & 73.98 \\
\textsc{RvG-Tree}       & 83.48 & 82.52 & 82.90 & 68.86 & 70.21 & 65.49 & 76.82 & 75.20 \\
\hline
\end{tabular}
\end{center}
\vspace{-2mm}
\caption{Top-1 Accuracy\% of ablative models on the three datasets with ground-truth object bounding boxes.}
\vspace{-2mm}
\label{table:gt_ablation}
\end{table*}
\begin{table*}[ht]
\footnotesize
\begin{center}
\begin{tabular}{| l | c | c | c || c | c | c || c | c |}
\hline
& \multicolumn{3}{c}{RefCOCO} & \multicolumn{3}{|c|}{RefCOCO+} & \multicolumn{2}{|c|}{RefCOCOg}\\
\cline{1-9}
& val & testA & testB & val & testA & testB & val & test\\
\hline\hline
Chain             & 72.01 & 75.55 & 67.34 & 59.44 & 62.78 & 53.90 & 64.05 & 64.73 \\
\textsc{RvG-Tree}-Fix        & 73.59 & 77.23 & 68.81 & 60.80 & 64.30 & 54.29 & 64.87 & 65.04\\
\textsc{RvG-Tree}-Scratch & 71.34 & 74.80 & 67.22 & 59.48 & 62.71 & 53.96 & 63.71 & 63.50 \\
\textsc{RvG-Tree}/Node  & 74.66 & 77.23 & 68.50 & 62.28 & 66.30 & 55.72 & 65.58 & 65.53  \\
\textsc{RvG-Tree}/S     & 74.22 & 76.89 & 68.02 & 61.95 & 65.77 & 55.01 & 65.12 & 64.99 \\
\textsc{RvG-Tree}/F     & 74.13 & 77.28 & 68.21 & 62.38 & 65.98 & 55.34 & 65.55 & 65.25 \\
\textsc{RvG-Tree}       & 75.06 & 78.61 & 69.85 & 63.51 & 67.45  & 56.66 & 66.95 & 66.51 \\
\hline
\end{tabular}
\end{center}
\vspace{-2mm}
\caption{Top-1 Accuracy\% of ablative models on the three datasets with detected object bounding boxes.}
\vspace{-2mm}
\label{table:det_ablation}
\end{table*}
\begin{table*}[t]
\begin{center}
\begin{tabular}{| l | c | c | c || c | c | c || c || c | c |}
\hline
 & \multicolumn{3}{c}{RefCOCO} & \multicolumn{3}{|c|}{RefCOCO+} & \multicolumn{3}{|c|}{RefCOCOg}\\
\cline{1-10}
 & val & testA & testB & val & testA & testB & val* & val & test\\
\hline\hline
MMI~\cite{mao2016generation}       & - & 63.15 & 64.21 & - & 48.73 & 42.13 & 62.14 & - & - \\
NegBag~\cite{nagaraja2016modeling} & 76.90 & 75.60 & 78.80 & - & - & - & - & - & 68.40 \\
Attribute~\cite{liu2017referring}    & - & 78.85 & 78.07 & - & 61.47 & 57.22 & 69.83 & - & - \\
CMN~\cite{hu2017modeling}  & - & 75.94 & 79.57 & - & 59.29 & 59.34 & 69.30 & - & - \\
VC~\cite{zhang2018grounding} & - & 78.98 & 82.39 & - & 62.56 & 62.90 & 73.98 & - & -\\
Speaker~\cite{yu2017joint}& 79.56 & 78.95 & 80.22 & 62.26 & 64.60 & 59.62 & 72.63 & 71.65 & 71.92 \\
Listener~\cite{yu2017joint}& 78.36 & 77.97 & 79.86 & 61.33 & 63.10 & 58.19 & 72.02 & 71.32 & 71.72\\
AccAttn~\cite{deng2018visual}& 81.27 & 81.17 & 80.01 & 65.56 & 68.76 & 60.63 & 73.18 & - & -\\
MAttN$^*$~\cite{yu2018mattnet} & 82.06 & 81.28 & \textbf{83.20} & 64.84 & 65.77 & 64.55 & - & 75.33 & 74.46 \\
\hline
\textsc{RvG-Tree} & 79.04 & 78.82 & 80.53 & 62.38 & 62.82 & 61.28 & 72.77 & 72.32 & 71.95 \\
\textsc{RvG-Tree}$^*$ & \textbf{83.48} & \textbf{82.52} & 82.90 & \textbf{68.86} & \textbf{70.21} & \textbf{65.49} & \textbf{76.29} & \textbf{76.82} & \textbf{75.20} \\
\hline
\end{tabular}
\end{center}
\vspace{-2mm}
\caption{Top-1 Accuracy\% of various grounding models on the three datasets with ground-truth object bounding boxes. $^*$ indicates that this model uses res101 features.}
\vspace{-2mm}
\label{table:gt_comp}
\end{table*}

\begin{table*}[t]
\begin{center}
\begin{tabular}{| l | c | c | c || c | c | c || c || c | c |}
\hline
 & \multicolumn{3}{c}{RefCOCO} & \multicolumn{3}{|c|}{RefCOCO+} & \multicolumn{3}{|c|}{RefCOCOg}\\
\cline{1-10}
 & val & testA & testB & val & testA & testB & val* & val & test\\
\hline\hline
MMI~\cite{mao2016generation} & - & 64.90 & 54.51 & - & 54.03 & 42.81 & 45.85 & - & - \\
NegBag~\cite{nagaraja2016modeling} & 57.30 & 58.60 & 56.40 & - & - & - & 39.50 & - & 49.50 \\
Attribute~\cite{liu2017referring} & - & 72.08 & 57.29 & - & 57.97 & 46.20 & 52.35 & - & - \\
CMN~\cite{hu2017modeling} & - & 71.03 & 65.77 & - & 54.32 & 47.76 & 57.47 & - & - \\
VC~\cite{zhang2018grounding} & - & 73.33 & 67.44 & - & 58.4 & 53.18 & 62.30 & - & - \\
Speaker~\cite{yu2017joint} & 69.48 & 72.95 & 63.43 & 55.71 & 60.43 & 48.74 & 59.51 & 60.21 & 59.63 \\
Listener~\cite{yu2017joint} & 68.95 & 72.95 & 62.98 & 54.89 & 59.61 & 48.44 & 58.32 & 59.33 & 59.21 \\
MAttN$^*$~\cite{yu2018mattnet} & 72.96 & 76.61 & 68.20 & 58.91 & 63.06 & 55.19 & - & 64.66 & 63.88 \\
\hline
\textsc{RvG-Tree} & 71.59 & 76.05 & 68.03 & 57.56 & 61.07 & 53.18 & 63.45 & 63.73 & 63.38 \\
\textsc{RvG-Tree}$^*$ & \textbf{75.06} & \textbf{78.61} & \textbf{69.85} & \textbf{63.51} & \textbf{67.45} & \textbf{56.66} & \textbf{66.20} & \textbf{66.95} & \textbf{66.51}\\
\hline
\end{tabular}%
\end{center}
\vspace{-2mm}
\caption{Top-1 Accuracy\% of various grounding models on the three datasets with detected object bounding boxes. $^*$ indicates that this model uses res101 features.}
\vspace{-4mm}
\label{table:det_comp}
\end{table*}

\item
\textbf{\textsc{RvG-Tree}/S}: This is the \textsc{RvG-Tree} model without accumulating the score form the score node. That is, we ignore the $S^{ls}$ in Eq.~\eqref{eq:3}.
\item
\textbf{\textsc{RvG-Tree}/F}: This is the \textsc{RvG-Tree} model without the node view pairwise score $S^{l}_p$ in Eq.~\eqref{eq:3}. This ablation discards the visual feature returned by the feature node. 
\end{itemize}

\begin{figure*}[t]
\begin{center}
\includegraphics[width=.95\linewidth]{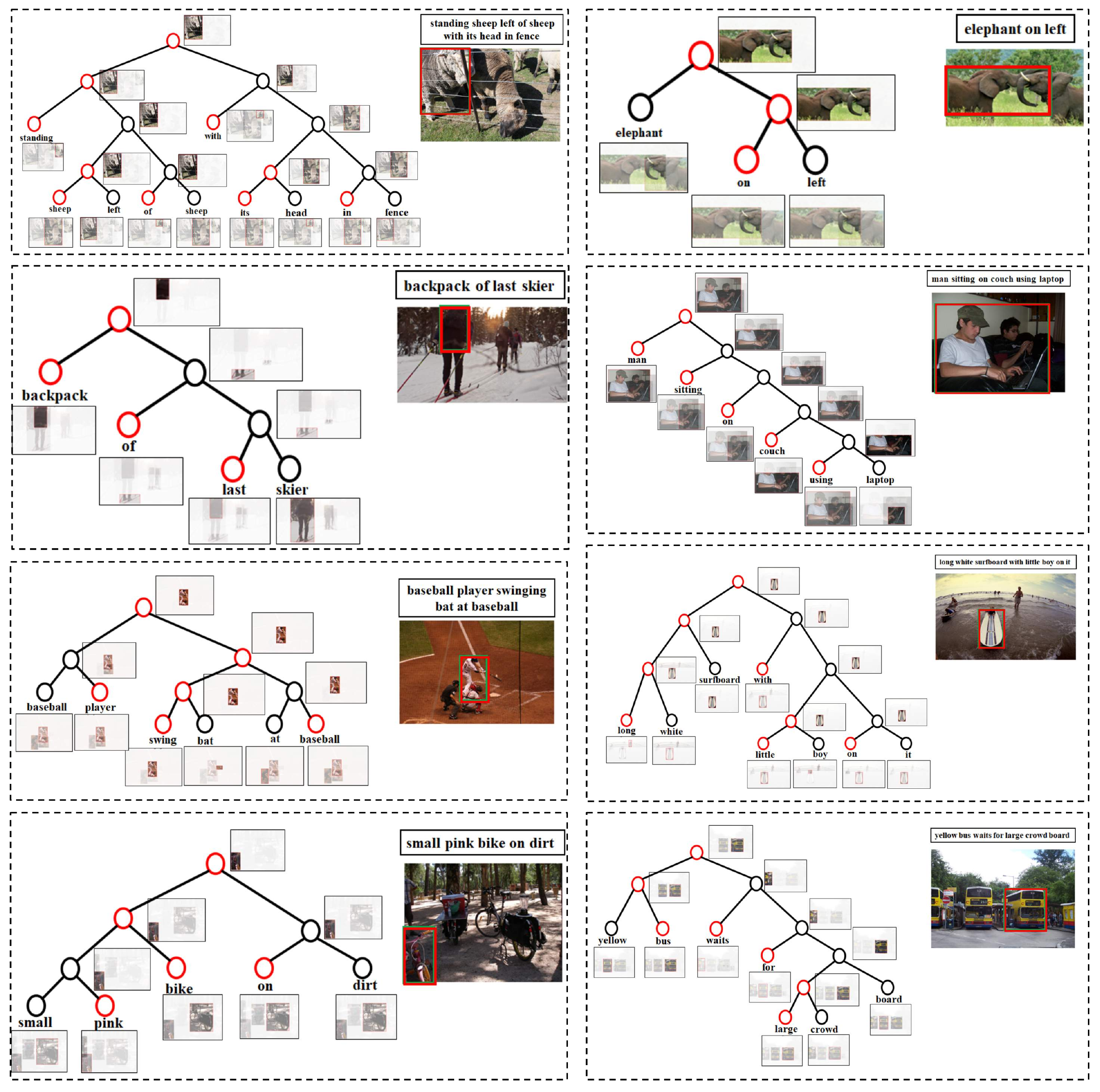}
\end{center}
\vspace{-4mm}
\caption{Qualitative results of correct grounding results from RefCOCOg. Red circle is the score node and black circle is the feature node. Each node is visualized by the score map or attention map of different regions: larger transparency indicates lower score. The ground-truth region is in green box and the result is in red box.}
\vspace{-4mm}
\label{fig:examples}
\end{figure*}

Table~\ref{table:gt_ablation} shows the grounding accuracies of the ablative methods on the three benchmarks. We can have the following observations:

1) On all datasets, \textsc{RvG-Tree}-Fix outperforms Chain. This is because that the tree structure is more suitable for language decomposition, that is, the tree sentence feature captures more structure semantics than the holistic language feature, especially for longer sentences such as RefCOCOg. However, we should note that the improvement is limited. We believe that this is due to that the \textsc{RvG-Tree}-Fix is essentially still a holistic language representation as only the root embedding is used in visual reasoning. This motivates us to design grounding score function that exploits the tree structure explicitly.

2) If we delete the in-node score, \textsc{RvG-Tree}/Node is significantly worse than the full \textsc{RvG-Tree}. The reasons are two-fold: first, the in-node score is an independent score to correct, if any, wrong grounding confidence passed from children nodes; second, the in-node score offers a more comprehensive linguistic view compared to its children view, as its related sub-sequence is a joint set of the two children.

3) The reason why \textsc{RvG-Tree}/S is worse than \textsc{RvG-Tree} is because the grounding score is not accumulated. This demonstrates that compositional visual reasoning is crucial for the task of grounding natural language.

4) Without the pairwise score calculated from the visual feature returned by the feature node, \textsc{RvG-Tree}/F is inferior to \textsc{RvG-Tree}. This demonstrates that the pairwise relationship is essential for distinguishing the referent from its context. This is especially useful for longer sentences in RefCOCOg: \textsc{RvG-Tree} considerably higher than its non-pairwise counterpart \textsc{RvG-Tree}/F.

5) We can see that tree construction from scratch generally fails. This is not surprising as it is quite challenging to learn language composition without any prior knowledge. Note that this observation also agrees with many works in reinforcement learning that requires supervised training as the teacher forcing~\cite{hu2017learning}.

\begin{figure*}[t]
\begin{center}
\includegraphics[width=.95\linewidth]{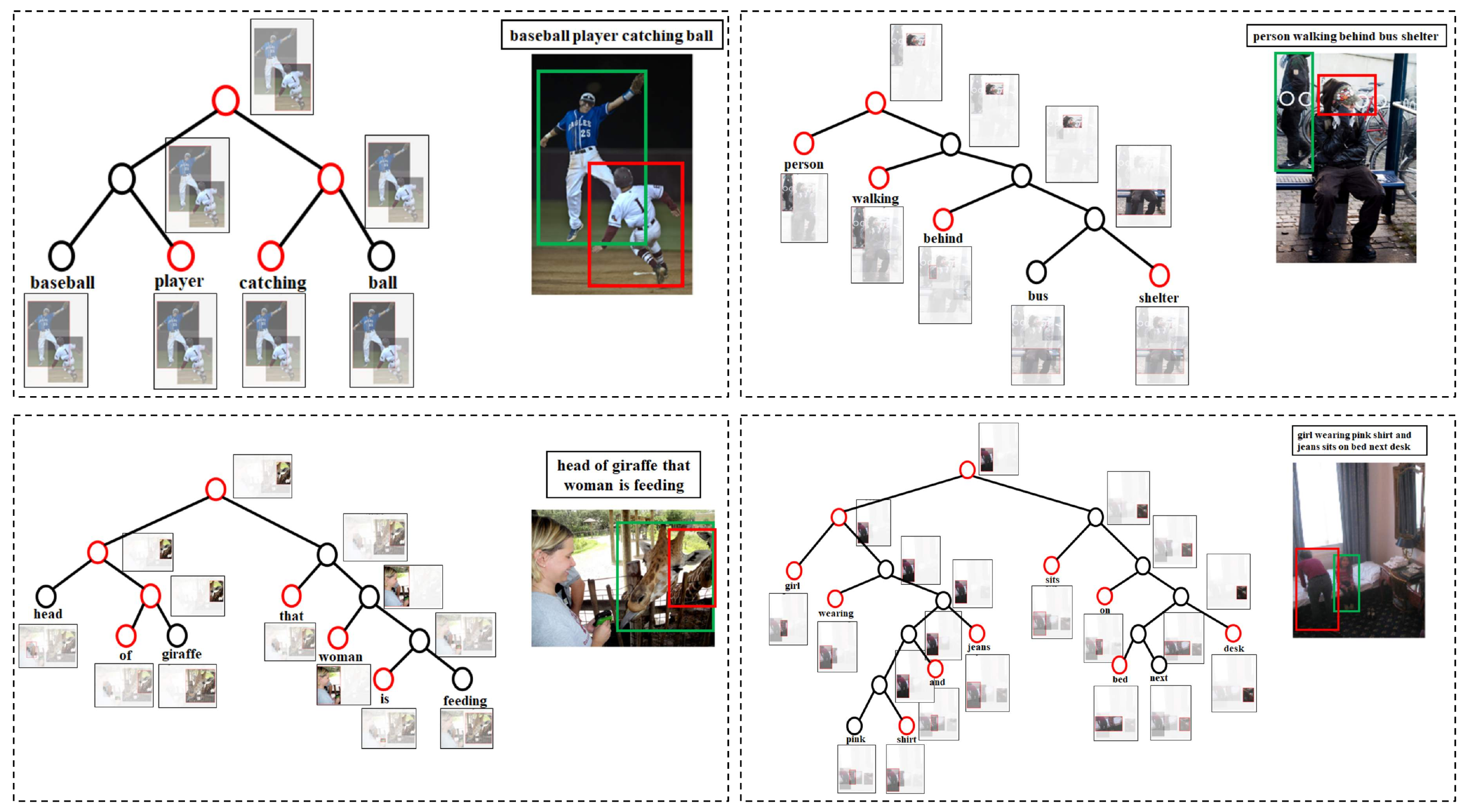}
\end{center}
\vspace{-4mm}
\caption{Qualitative results of incorrect grounding results from RefCOCOg. Red circle is the score node and black circle is the feature node. Each node is visualized by the score map or attention map of different regions: larger transparency indicates lower score. The ground-truth region is in green box and the result is in red box.}
\vspace{-4mm}
\label{fig:examples2}
\end{figure*}

\vspace{-3mm}
\subsection{Comparison with State-of-the-Arts}
We compared \textsc{RvG-Tree} with state-of-the-art referring expression grounding models published in recent years. In light of whether the model requires language composition, these comparing methods can be categorized as follows: 1) the resultant region is the one that can generate the referring expression sentence with maximum probability (by  maximizing a posteriori probability), such as the pioneering \textbf{MMI}~\cite{mao2016generation}, \textbf{Attribute}~\cite{liu2017referring}, and the \textbf{Speaker}~\cite{yu2017joint} and \textsc{Listener}~\cite{yu2017joint}. Though the score mechanism exploits language composition during the sentence generation, it does not consider the different visual regions while generation. 2) localization based grounding methods that only use holistic language features, such as \textbf{NegBag}~\cite{nagaraja2016modeling}. 3) localization based grounding methods that use language composition such as \textbf{CMN}~\cite{hu2017modeling}, \textbf{VC}~\cite{zhang2018grounding}, \textbf{MAttN}~\cite{yu2018mattnet}. Note that \textsc{RvG-Tree} belongs to the family of the last localization based grounding models, but its language composition is much more fine-grained than the previous state-of-the-arts. 
\subsubsection{Results on Ground-Truth Regions}
From the results on RefCOCO, RefCOCO+, and RefCOCOg in Table~\ref{table:gt_comp}, we can see that \textsc{RvG-Tree} achieves the state-of-the-art performance. We believe that the improvement is attributed to the recursive grounding score along the tree structure. First, on all datasets, \textsc{RvG-Tree} outperforms all the other sentence generation-comprehension methods: MMI, Attribute, Speaker, Listener, that do not consider language structure and visual context. Second, \textsc{RvG-Tree} outperforms the triplet compositional models such as CMN and VC. The improvement is attributed to the fact that \textsc{RvG-Tree} recursively applies the triplet-like grounding score along the tree structure, and hence the grounding confidence is more comprehensive than those methods. This demonstrates the effectiveness of the recursive fashion of the compositional grounding.

As shown in Fig.~\ref{fig:examples}, we illustrate some qualitative correct grounding results with the learned tree structure and their intermediate grounding process. Each intermediate node is visualized by the soft-attention map for the contextual feature if it is a feature node, or the score map of regions if it is a score node. We can see that most of the grounding results show reasonable intermediate results. For example, in the tree of ``baseball player swinging bat at baseball'', ``baseball player'' on the left sub-tree scores high value for both of the player regions; however, by the help of the right sub-tree ``swinging bat at baseball'' which scores higher for the person who is ``swinging'', \textsc{RvG-Tree} can pinpoint the correct referent in the end. This is intuitively similar to human reasoning and the purpose of using attributive clause in English. Moreover, the classification of score and feature node also sheds some light on the tree structure. For example, due to that the goal is to distinguish the baseball player, the ``baseball player'' itself is working as a context feature vector, \ie, feature node, and the score should be thus dominated by the score node with words ``swinging bat at baseball''. If the image contains more than one object class, the top score node is usually connected with the referent, \eg, ``backpack'' and ``bike''. We also note that some of the tree structure is trivially deep, by always merging the last two nodes, for example, ``man sitting on couch using laptop'' and ``backpack of last skier''. So far, we cannot find out what the cause is, but it seems that such trivial structures are still reasonable for their corresponding grounding scenario.

Fig.~\ref{fig:examples2} shows some failure cases of \textsc{RvG-Tree} performed on RefCOCOg. We can see that most of the errors are caused by the wrong comprehension of the nuanced visual relationships, especially those contain comparative semantics. For the example of ``girl wearing pink shirt and jeans sits on bed next desk'', from the intermediate results, we are happy to see that our model correctly grounds ``pink shirt'' and ``jeans'' but it fails to distinguish the which girl is ``on'' bed and ``next'' to the desk. In fact, both of the girls are ``on'' and ``next'' visually, that is, our visual model does not fail. However, the true semantic meaning should be ``sits on'', which is very challenging for current model to discover. This failure may shed some lights on our future direction. As another example, though the model successfully identifies the ``walking'' out of ``person walking behind bus shelter'', its final score is lower (though close) than something that is ``behind''. This reveals some the drawback of \textsc{RvG-Tree} is that more linguistic prior knowledge should be used. For example, if we can identify the referent adjective is ``walking'', we may assign higher weights on the ``walking person'' but not the background ``behind''. Some failures are due to the imperfect visual recognition models for human actions, for example, ``catching''. This kind of failures is expected to be resolved by more fine-grained human action recognizer using parsed human bodies. 

\subsubsection{Results on Detected Regions}
So far, our results are on grounding tasks with ground-truth object bounding boxes, that is, each visual region is guaranteed to be a valid object. Though this setting eliminates the errors from imperfect object detection and hence allows the algorithmic design to focus on the discrimination between similar regions, it is not practical in real applications, where we always use object detectors to obtain the image regions. Therefore, it is also necessary to evaluate our methods on the various number of detected bounding boxes using Faster-RCNN. We followed the classic MS-COCO detection task NMS to obtain 10 to 100 objects per image. 

From Table~\ref{table:det_comp}, we can see that the performance of all methods has been dropped due to the imperfect bounding box detections. However, the observations discovered in previous ground-truth box experiments still hold in detected boxes. It is worth noting that the performance of methods without compositional reasoning: MMI, NegBag, Attribute, Speaker, and Listener, dropped the most. It is due to the fact that these models only learn language and region associations that are easily overfitted to the distribution bias of the ground-truth bounding boxes; when our test moves to the detected boxes, the bias no longer holds. Compared to other compositional reasoning models: CMN, VC, and MAttN, our \textsc{RvG-Tree} is better.  This demonstrates the robustness of our model to noisy visual regions. 

\section{Conclusions}
In this paper, we propose a novel model called Recursive Grounding Tree (\textsc{RvG-Tree}) that localizes the target region by recursively accumulating the vision-language grounding scores. To the best of our knowledge, this is the first visual grounding model that leverages the entire language compositions. \textsc{RvG-Tree} learns to compose a binary tree structure, with proper supervised pre-training, and the final grounding score of the root is defined in a recursive way that accumulates grounding confidence from two children sub-trees. This process is fully differentiable. We conducted extensive ablative, quantitative, and qualitative experiments on three benchmark datasets of referring expression grounding. Results demonstrate the effectiveness of \textsc{RvG-Tree} in visual reasoning: 1) the complex language composition is decomposed into easier sub-grounding tasks, and 2) the overall grounding score can be easily explained by inspecting the intermediate grounding results along the tree. Therefore, compared to existing models, \textsc{RvG-Tree} is more compositional and explainable.

The key limitation of \textsc{RvG-Tree} is that the linguistic prior knowledge is not fully exploited. However, acquiring such knowledge is essentially important for machine comprehension of natural language and then grounding it based on the comprehension. Although we have demonstrated that using constituency parsers as prior knowledge can boost the performance, it is not sufficient. In future, we are going to add more linguistic cues into the recursive grounding score function to guide the confidence accumulation.
\ifCLASSOPTIONcompsoc
  \section*{Acknowledgments}
\else
  \section*{Acknowledgment}
\fi

We thank the editors and the reviewers for their helpful suggestions.
This work was supported by the National Key Research and Development Program under Grant 2017YFB1002203, the National Natural Science Foundation of China under Grant 61722204 and 61732007, and Alibaba-NTU Singapore Joint Research Institute.

\ifCLASSOPTIONcaptionsoff
  \newpage
\fi
\bibliography{citation}
\bibliographystyle{IEEEtran}

\begin{IEEEbiography}[{\includegraphics[width=1in,height=1.25in,clip,keepaspectratio]{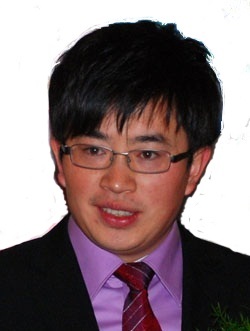}}]
{Richang Hong} (M’11) received the Ph.D. degree from the University of Science and Technology of China, Hefei, China, in 2008. He is currently a Professor with the Hefei University of Technology, Hefei. His research interests include multimedia content analysis and social media, in which he has co-authored over 100 publications. He is a member of the ACM and an Executive Committee Member of the ACM SIGMM China Chapter. He was a recipient of the Best Paper Award at the ACM Multimedia 2010, the Best Paper Award at the ACM ICMR 2015, and the Honorable Mention of the IEEE TRANSACTIONS ON MULTIMEDIA Best Paper Award 2015. He was an Associate Editor of the IEEE Multimedia Magazine, Information Sciences and Signal Processing, Elsevier, and the Technical Program Chair of the MMM 2016, ICIMCS 2017, and PCM 2018.
\end{IEEEbiography}

\begin{IEEEbiography}[{\includegraphics[width=1in,height=1.25in,clip,keepaspectratio]{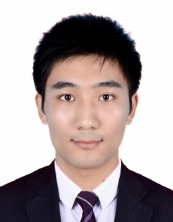}}]
{Daqing Liu} received the B.E. degree in Automation from Chang'an University,
Xi'an, China, in 2016, and currently working
toward the Ph.D. degree from the Department of Automation, University of Science and Technology of
China, Hefei, China. His research interests mainly include computer vision and multimedia.
\end{IEEEbiography}

\begin{IEEEbiography}[{\includegraphics[width=1in,height=1.25in,clip,keepaspectratio]{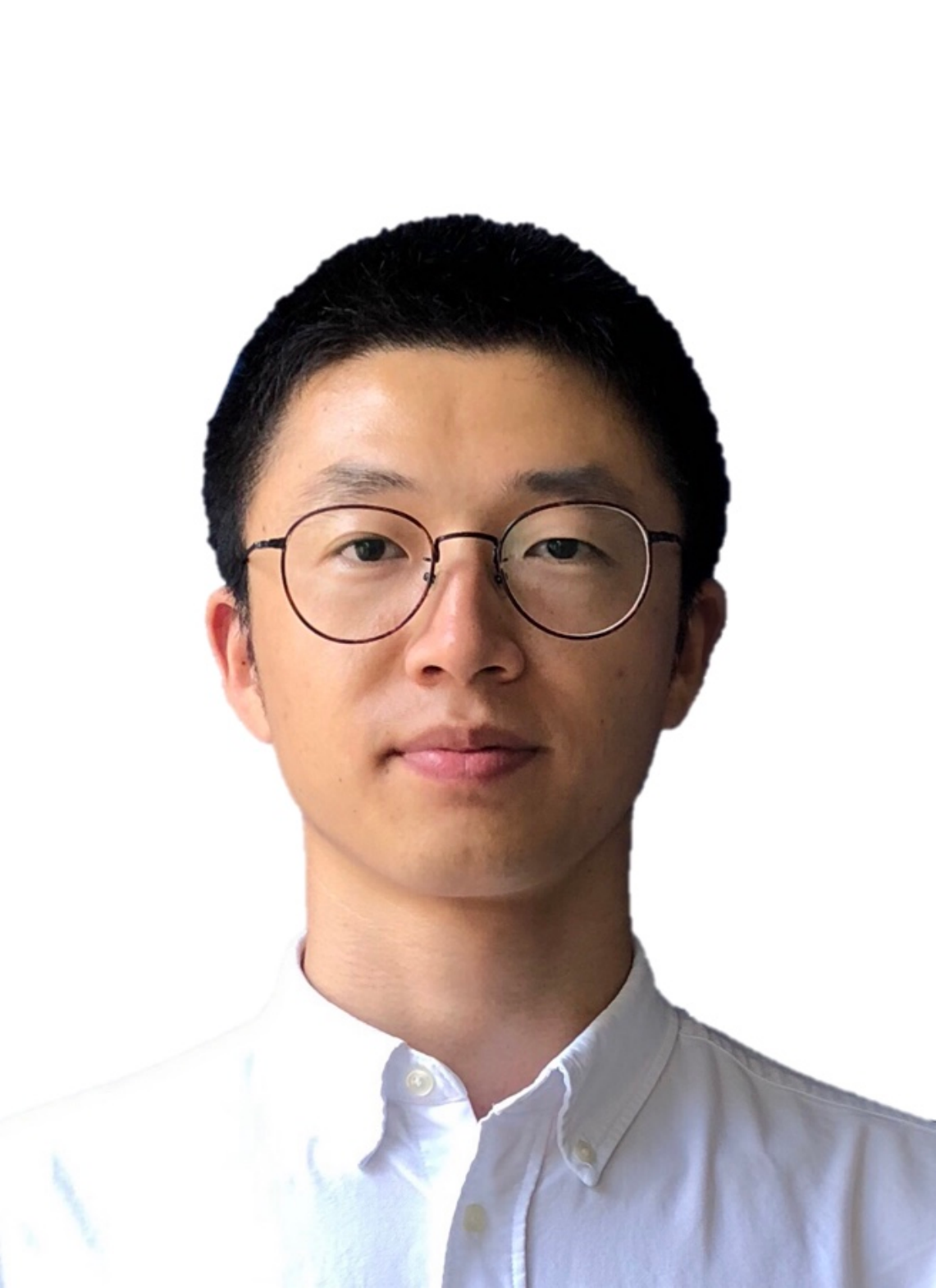}}]
{Xiaoyu Mo} received the B.E. degree from Yangzhou University, Yangzhou, China, in 2015, the M.E. degree from Huazhong University of Science and Technology, Wuhan, China, in 2017. He was a research associate with Nanyang Technological University, Singapore,  from Sep. 2017 to Jan. 2019. He is now a Ph.D. student in Nanyang Technological University. His research interests include consensus and autonomous vehicles.
\end{IEEEbiography}

\begin{IEEEbiography}[{\includegraphics[width=1in,height=1.25in,clip,keepaspectratio]{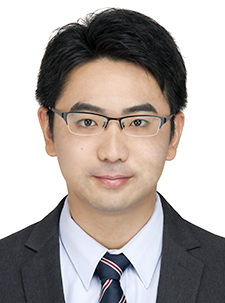}}]
{Xiangnan He} is currently a professor with the University of Science and Technology of China (USTC). He received his Ph.D. in Computer Science from National University of Singapore (NUS) in 2016, and did postdoctoral research in NUS until 2018. His research interests span information retrieval, data mining, and multi-media analytics. He has over 50 publications appeared in several top conferences such as SIGIR, WWW, and MM, and journals including TKDE, TOIS, and TMM. His work on recommender systems has received the Best Paper Award Honourable Mention in WWW 2018 and ACM SIGIR 2016. Moreover, he has served as the PC member for several top conferences including SIGIR, WWW, MM, KDD etc., and the regular reviewer for journals including TKDE, TOIS, TMM, TNNLS etc.
\end{IEEEbiography}

\begin{IEEEbiography}[{\includegraphics[width=1in,height=1.25in,clip,keepaspectratio]{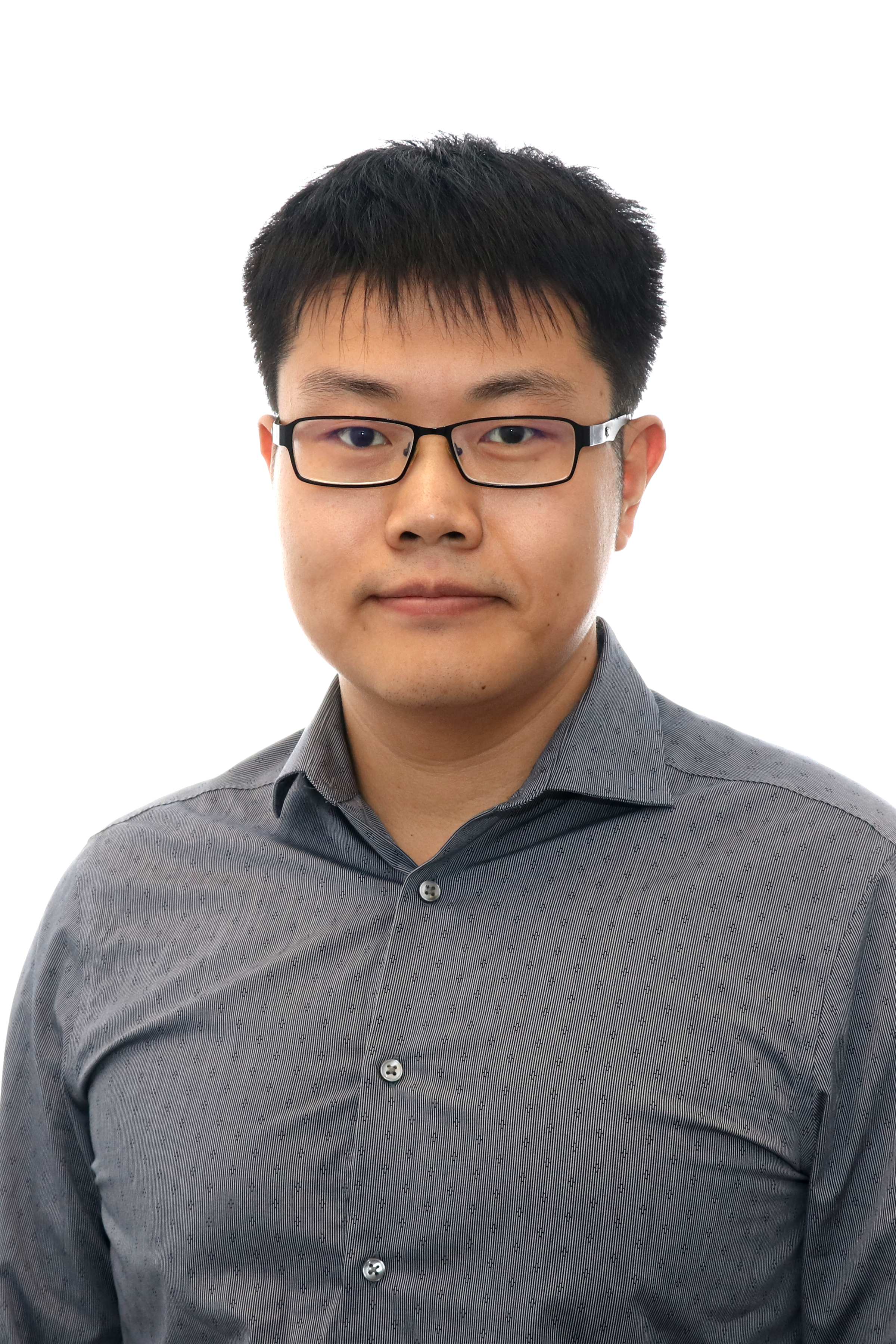}}]
{Hanwang Zhang} is currently an Assistant Professor at Nanyang Technological University, Singapore. He was a research scientist at the Department of Computer Science, Columbia University, USA. He has received the B.Eng (Hons.) degree in computer science from Zhejiang University, Hangzhou, China, in 2009, and the Ph.D. degree in computer science from the National University of Singapore in 2014. His research interest includes computer vision, multimedia, and social media. Dr. Zhang is the recipient of the Best Demo runner-up award in ACM MM 2012, the Best Student Paper award in ACM MM 2013, and the Best Paper Honorable Mention in ACM SIGIR 2016，and TOMM best paper award 2018. He is also the winner of Best Ph.D. Thesis Award of School of Computing, National University of Singapore, 2014.
\end{IEEEbiography}



\end{document}